\PassOptionsToPackage{dvipsnames,table}{xcolor}
\documentclass{article}

\usepackage{microtype}
\usepackage{graphicx}
\usepackage{subcaption}
\usepackage{booktabs} 

\usepackage{hyperref}

\usepackage[preprint]{icml2026}

\usepackage{amsmath}
\usepackage{amssymb}
\usepackage{mathtools}
\usepackage{amsthm}
\usepackage{multirow}
\usepackage{makecell}
\usepackage{colortbl}
\usepackage{tabularx}
\usepackage{titletoc}
\usepackage{fontawesome5}
\usepackage[breakable]{tcolorbox}
\usepackage{colortbl}
\usepackage{enumitem}
\definecolor{myblue}{RGB}{216,236,255}
\newcommand{\down}[1]{\textcolor{Maroon}{\small\ $\downarrow$ {#1}}}
\newcommand{\uprise}[1]{\textcolor{OliveGreen}{\small\ $\uparrow$ {#1}}}
\newcommand{\basex}[1]{\textcolor{gray!50}{\small\ $\uparrow$ {#1}}}

\usepackage[capitalize,noabbrev]{cleveref}

\theoremstyle{plain}
\newtheorem{theorem}{Theorem}[section]
\newtheorem{proposition}[theorem]{Proposition}

\theoremstyle{definition}
\newtheorem{definition}[theorem]{Definition}

\theoremstyle{remark}

\usepackage[textsize=tiny]{todonotes}

\icmltitlerunning{Replay Failures as Successes: Sample-Efficient Reinforcement Learning for Instruction Following}

\begin{document}

\twocolumn[
\icmltitle{Replay Failures as Successes:\\Sample-Efficient Reinforcement Learning for Instruction Following}



  \icmlsetsymbol{equal}{*}

  \begin{icmlauthorlist}
    \icmlauthor{Kongcheng Zhang}{zju}
    \icmlauthor{Qi Yao}{cainiao}
    \icmlauthor{Shunyu Liu}{ntu,equal}
    \icmlauthor{Wenjian Zhang}{dut}
    \icmlauthor{Min Cen}{ustc}
    \icmlauthor{Yang Zhou}{zju} \\
    \icmlauthor{Wenkai Fang}{zju}
    \icmlauthor{Yiru Zhao}{alibaba}
    \icmlauthor{Baisheng Lai}{cnic}
    \icmlauthor{Mingli Song}{zju}
  \end{icmlauthorlist}

  \icmlaffiliation{zju}{Zhejiang University}
  \icmlaffiliation{cainiao}{Cainiao Network}
  \icmlaffiliation{ntu}{Nanyang Technological University}
  \icmlaffiliation{dut}{Dalian University of Technology}
  \icmlaffiliation{ustc}{University of Science and Technology of China}
  \icmlaffiliation{alibaba}{Alibaba Cloud Computing}
  \icmlaffiliation{cnic}{Chinese Academy of Sciences}
  \centering \textsuperscript{1}Zhejiang University, \textsuperscript{2}Cainiao Network,  \textsuperscript{3}Nanyang Technological University,
  \textsuperscript{4}Dalian University of Technology \\
  \centering \textsuperscript{5}University of Science and Technology of China, \textsuperscript{6}Alibaba Cloud Computing, \textsuperscript{7}Chinese Academy of Sciences \\
  \centering \texttt{zhangkc@zju.edu.cn} ,\texttt{yq223369@alibaba-inc.com}, \texttt{shunyu.liu.cs@gmail.com}

  \icmlkeywords{Machine Learning, ICML}

  \vskip 0.3in
]



\renewcommand{\thefootnote}{} 
\footnotetext{* Corresponding author}
\renewcommand{\thefootnote}{} 
\printAffiliationsAndNotice{}  

\begin{abstract}
Reinforcement Learning~(RL) has shown promise for aligning Large Language Models~(LLMs) to follow instructions with various constraints. Despite the encouraging results, RL improvement inevitably relies on sampling successful, high-quality responses; however, the initial model often struggles to generate responses that satisfy all constraints due to its limited capabilities, yielding sparse or indistinguishable rewards that impede learning. In this work, we propose \textit{\textbf{H}indsight \textbf{i}nstruction \textbf{R}eplay}~(HiR), a novel sample-efficient RL framework for complex instruction following tasks, which employs a \textit{select}-then-\textit{rewrite} strategy to \textit{replay failed attempts as successes} based on the constraints that have been satisfied in hindsight. We perform RL on these replayed samples as well as the original ones, theoretically framing the objective as dual-preference learning at both the instruction- and response-level to enable efficient optimization using only a binary reward signal. Extensive experiments demonstrate that the proposed HiR yields promising results across different instruction following tasks, while requiring less computational budget.

\begin{center}
\faGithub \, \href{https://github.com/sastpg/HIR}{Code} \quad \quad \includegraphics[height=2.1ex]{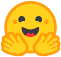} \,\href{https://huggingface.co/datasets/sastpg/HIR-16K}{Dataset}
\end{center}
\end{abstract}

\begin{figure*}[!t]
    \centering
    \begin{minipage}{0.5\textwidth}
        \centering
        \includegraphics[width=\linewidth]{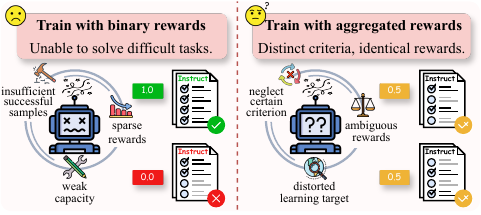}
    \end{minipage}
    \hspace{-0.02\textwidth}
    \begin{minipage}{0.5\textwidth}
        \centering
        \includegraphics[width=\linewidth]{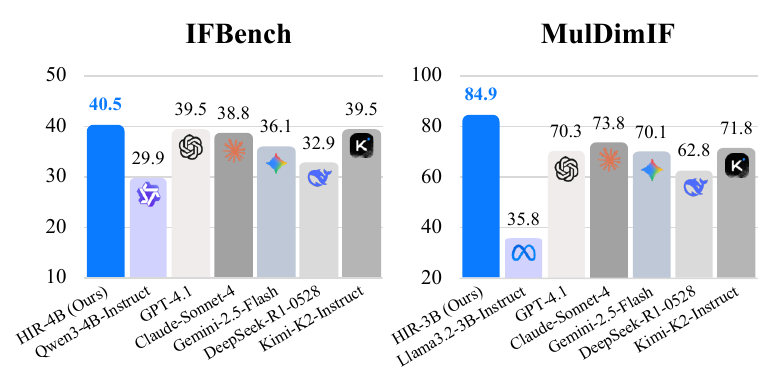}
    \end{minipage}
    \vspace{-5pt}
    \caption{(Left) A conceptual illustration of the sparse and indistinguishable reward problem in current RLVR methods for instruction following tasks. (Right) Performance comparison between small LLMs trained by HiR and frontier LLMs on different benchmarks.}
    \label{fig:drawback_acc}
    \vspace{-5pt}
\end{figure*}

\section{Introduction}
Large Language Models (LLMs) have demonstrated remarkable capabilities across a wide spectrum of natural language tasks, such as content creation~\citep{minaee2024large, qian2023creator, lee2023making}, financial analysis~\citep{arun2023numerical, kim2024financial}, and robotic control~\citep{krause2023palm, firoozi2025foundation, huang2025graphcot}. Among these capabilities, instruction following has attracted substantial attention, driven by the growing reliance of intelligent applications on LLMs~\citep{zhou2023characterglm, li2025websailor, qiao2025webresearcher} to reliably interpret user intent and perform specific tasks. However, real-world instructions typically involve diverse, multiple constraints, ranging from output formatting to logical consistency, which makes it challenging for LLMs to 
satisfy all requirements at the same time~\citep{lior2025wildifeval, qi2025agentif}.

Recent breakthroughs in Reinforcement Learning with Verifiable Rewards (RLVR)~\citep{lambert2024tulu, guo2025deepseek, zhang2025survey} have provided a promising strategy to incentivize sophisticated reasoning patterns via rule-based rewards. Despite the leading results in mathematical analysis~\citep{zhang2025reasoning, zeng2025simplerl} and algorithmic programming~\citep{zhu2025codev}, the application of RL remains underexplored in open-ended tasks like complex instruction following~\citep{wen2024benchmarking, sakai2025revisiting, song2025ifir, ye2025multi, wang2025ask}, where straightforward ground-truth labels are often unavailable. To bridge this gap, several recent works~\citep{lambert2024tulu, peng2025verif, qin2025incentivizing} adopt the ``LLM-as-a-Judge" paradigm, in which a powerful judge model assigns reward signals by scoring model responses against evaluable criteria derived from the instructions.

However, a critical bottleneck remains as RL relies on self-exploration to improve, yet the initial model may struggle to generate responses that satisfy all given constraints due to its limited capabilities, even after many attempts~\citep{yue2025does, wu2025invisible}. As a result, the learning signal becomes highly sparse when using binary rewards~\citep{peng2025verif}, \textit{i.e.}, a response is rewarded only if it perfectly meets every constraint. To mitigate this sparsity, prior works~\citep{Pyatkin2025GeneralizingVI, qin2025incentivizing} often adopt an aggregated reward signal, averaging individual scores for each constraint to provide a denser signal. Although this aggregated mechanism can stabilize training, it poses a risk of reward ambiguity. As shown in the left part of Figure~\ref{fig:drawback_acc}, two responses could share the same aggregated reward while exhibiting substantial variation in adherence to constraints, which obscures the underlying causes of failures. Worse still, this ambiguity may distort the intended learning goals: treating responses with higher rewards as preferable could misguide the model to neglect certain constraints, since both high-reward and low-reward responses may have aspects where they outperform the other.

To tackle these issues, we propose \textit{\textbf{H}indsight \textbf{i}nstruction \textbf{R}eplay} (HiR), a sample-efficient RL framework that employs a \textit{select}-then-\textit{rewrite} replay strategy to solve multi-constraint instruction following tasks. Technically, we first select valuable failure samples in a curriculum-based manner, prioritizing response diversity and then gradually weighing constraint integrity as training proceeds. This trade-off dynamically accounts for the varying contribution of each sample across different learning stages, thereby improving both generalization ability and learning quality. Next, the instructions of selected samples are rewritten into ``hindsight" pseudo-instructions by removing unmet constraints, followed by assigning positive rewards on these samples for replay. Finally, we perform RL on both original and replayed samples, enabling efficient learning with only a binary reward signal. The theoretical analysis reveals that our training objective not only aligns response preferences but also captures nuanced differences among instructions, facilitating the model to explicitly identify specific unmet constraints instead of relying on ambiguous rewards. Our key contributions are summarized as follows:

\begin{itemize}[leftmargin=*]
\item We propose HiR as a novel paradigm in RL for instruction following tasks, which pioneers the transition of failure responses into successful ones by constructing hindsight pseudo-instructions, thereby providing more informative learning signals to enable efficient optimization.
\item We introduce a \textit{select}-then-\textit{rewrite} replay strategy that considers both response diversity and constraint integrity, complemented by a curriculum schedule to balance the exploration-exploitation trade-off during training.
\item Extensive experiments demonstrate that HiR yields results superior to existing counterparts with even less computational budget. Notably, HiR enables small LLMs to achieve performance on par with leading LLMs, as shown in the right part of Figure~\ref{fig:drawback_acc}.
\end{itemize}

\section{Background and Notation}
\subsection{Instruction Following}
Our goal is to enhance the capability of LLMs in following complex instructions. We now formally define the instruction following task. Let an instruction \( q \) consists of a task description \( x \) and a set of constraints \( \mathcal{C} = \{c_1, c_2, \dots, c_n\} \). Following the formulation of \citet{zhou2023instruction}, an LLM parameterized by \( \theta \) is considered as following the instruction if its output \( y \) adhere to all specified constraints in \( \mathcal{C} \). We further categorize the constraint set \( \mathcal{C} \) into two types inspired by~\citet{peng2025verif}: \textit{Hard constraints} that are verifiable via deterministic rules or code (\textit{e.g.}, length and format); \textit{Soft constraints} requiring semantic evaluation (\textit{e.g.}, style or coherence). To verify whether a response meets these constraints, we adopt a hybrid evaluation approach: hard constraints are assessed using rule-based verifiers, while soft constraints are evaluated via the LLM-as-a-judge mechanism~\citep{li2025generation}. The evaluation prompt for the judge LLM is presented in Appendix~\ref{app:prompt}. This hybrid methodology enables efficient and comprehensive evaluation of instruction adherence.

\begin{figure*}
    \centering
    \includegraphics[width=\linewidth]{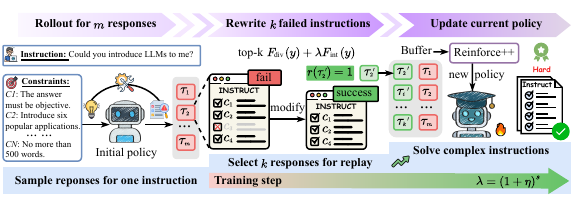}
    \vspace{-18pt}
    \caption{The overall framework of HiR with a \textit{select}-then-\textit{rewrite} replay strategy. First, we generate samples and select valuable failure attempts for replay with a curriculum schedule. Then we rewrite the instructions of selected samples into ``hindsight'' pseudo-instructions by removing the unmet constraints. Finally, we perform RL on both replayed samples as well as the original ones.}
    \vspace{-10pt}
    \label{fig:framework}
\end{figure*}

\subsection{Evaluation Metrics}
For a single constraint, we use a binary function (0 or 1) $\mathbb{I}(q, y, c_i)$ to indicate whether a response $y$ meets the constraint $c_i$ (true or false):
\begin{equation}
\mathbb{I}(q, y, c_i) = 
\begin{cases}
\text{Rule}(c_i, y), & \text{if } c_i \in \mathcal{C}_{\text{hard}}, \\
\text{LLM}(c_i, y), & \text{if } c_i \in \mathcal{C}_{\text{soft}},
\end{cases}
\end{equation}
where \( \mathcal{C}_{\text{hard}} \) and \( \mathcal{C}_{\text{soft}} \) denote the sets of hard and soft constraints, respectively. Extending this to the full constraints, we introduce two metrics at different granularities to measure performance in the following.

\textbf{Instruction-Level Accuracy (ILA)}. This metric reflects strict adherence to the entire instruction, where a response $y$ is considered correct only if it satisfies \textit{every} constraint associated with the instruction $q$:
\begin{equation}
\label{eq:pla}
\text{ILA}(q, y, \mathcal{C}) = \prod_{c_i \in \mathcal{C}} \mathbb{I}(q, y, c_i).
\end{equation}
\textbf{Constraint-Level Accuracy (CLA)}. This metric measures the ability to follow individual atomic constraints, which is calculated as the percentage of satisfied constraints:
\begin{equation}
\text{CLA}(q, y, \mathcal{C}) = \frac{1}{|\mathcal{C}|} \sum_{c_i \in \mathcal{C}} \mathbb{I}(q, y, c_i),
\end{equation}
When employing ILA as the reward signal for RL training, it often leads to sparse reward problem. Although CLA can provide a granular signal, it still suffers from reward ambiguity, as illustrated in the left side of Figure~\ref{fig:drawback_acc}.

\section{Hindsight Instruction Replay}
During rollout generation on instructions with multiple constraints, LLMs typically fail to generate sufficient perfect responses for training, especially for models with weaker capabilities. The core idea behind our method is to learn from failures by replaying failed attempts under hindsight pseudo-instructions: despite these samples may not help models learn how to fully satisfy original instructions, they definitely tell something about how to deal with partial constraints. In what follows, we introduce our sample-efficient RL framework HiR with a \textit{select}-then-\textit{rewrite} replay strategy as illustrated in Figure~\ref{fig:framework}.

\subsection{\textit{Select}-then-\textit{Rewrite} Replay Strategy}
\begin{algorithm}[t]
\caption{\textsc{Select-Rewrite}($\mathcal{G}$, $k$)}
\label{alg:selection}
\begin{algorithmic}[1]
\STATE \textbf{Input:} Sampling group $\mathcal{G}$, $k$
\STATE Initialize $\mathcal{T} \leftarrow \emptyset, \mathcal{H} \leftarrow \emptyset$
\FOR{each tuple ($q_i$, $y_i$, $\mathcal{C}$) in group $\mathcal{G}$}
    \STATE  Calculate score $F(y_i) = \lambda F_{int}(y_i) + F_{div}(y_i)$
\ENDFOR
\STATE Add to $\mathcal{T}$ tuples with top-$k$ score $F(y_i)$ \textcolor{OliveGreen}{\textit{// Select}}
\FOR{each tuple ($q_i$, $y_i$, $\mathcal{C}$) in $\mathcal{T}$}
    \STATE Identify satisfied constraints, \textit{i.e.,} \\ $\mathcal{C}^{'}_{i} \leftarrow \{c \in \mathcal{C}_i \mid \mathbb{I}(q_i, y_i, c) = 1\}$
    \STATE Rewrite instruction $q_i$ as $q^{'}_{i}$ using $\mathcal{C}^{'}_{i}$ \textcolor{OliveGreen}{\textit{// Rewrite}}
    \STATE Add tuple ($q_{i}^{'}$, $y_i$, $\mathcal{C}^{'}_{i}$) to $\mathcal{H}$
\ENDFOR
\STATE \textbf{Return:} Hindsight replay buffer $\mathcal{H}$ with size $k$
\end{algorithmic}
\end{algorithm}
Although replaying all partially failed attempts is possible, not all of them are equally informative to different learning stages. Samples deviating too far from the original constraints provide limited guidance toward following the targeted instructions; while some exhibit high similarity, thus redundant for learning. Consequently, different samples may contribute unevenly to the desired target. Recent studies~\citep{hammoud2025train, xie2025logic} have shown that a well-designed curriculum learning approach in RL for LLMs can always improve the final performance and learning efficiency. Motivated by this, we employ a selection criterion to replay a subset of failed responses $\mathcal{T}$ from each sampling group $\mathcal{G}$ based on the scheduled \textit{response diversity} and \textit{constraint integrity}. Specifically, we prefer more diversity at the early training stage and gradually increase the weight on constraint integrity in our selection strategy as training proceeds, which can be formulated as the following function over the subset $\mathcal{T}$ with size $k$:
\begin{equation}
\label{eq:selection}
\mathcal{T} \triangleq \operatorname*{arg\,max}_{\mathcal{T} \subseteq \mathcal{G},\, |\mathcal{T}|=k} \sum_{y \in \mathcal{T}} \big(F_{div}(y) +\lambda F_{int}(y) \big).
\end{equation}
Under the formulation in Eq. (\ref{eq:selection}), the optimal subset $\mathcal{T}$ is obtained by selecting the top-$k$ responses according to the score $F(y) = F_{div}(y) +\lambda F_{int}(y)$.

The first term $F_{div}(y)$ measures the diversity of the response. We use the response entropy to compute $F_{div}(y)$:
\begin{equation}
F_{div}(y) = -\sum_{t=1}^T \sum_{j=1}^V p_{t, j} \log p_{t, j},
\end{equation}
where $(p_{t, 1}, p_{t, 2}, ..., p_{t, V}) \sim \pi_\theta( \cdot | q, y_{<t})$ denote the corresponding probability distribution of $t$-th token over model vocabulary, $V$ denotes the vocabulary size, and $T$ denotes the token length of response $y$.

The second term $F_{int}(y)$, associated with a curriculum weight $\lambda$, reflects the integrity of original constraints. It is calculated by the percentage of satisfied constraints:
\begin{equation}
\label{eq:proximity}
F_{int}(y) = \frac{1}{|\mathcal{C}|} \sum_{c_i \in \mathcal{C}} \mathbb{I}({q, y, c_i}).
\end{equation}
Intuitively, the transition from response diversity to constraint integrity in our selection strategy reflects the classical exploitation–exploration trade-off. At early training stages, replaying trajectories with higher entropy encourages the model to explore uncertain yet informative patterns. However, the emphasis on diversity in later stages can distract learning, since the model has sufficiently explored the solution space and it becomes more important to focus on learning how to achieve all desired constraints of an instruction. We implement this transition by gradually increasing the weight $\lambda$ on constraint integrity during training: 
\begin{equation}
\label{eq:lambda}
    \lambda = (1+\eta)^s \cdot \lambda_0,
\end{equation}
where $\eta \in [0,1]$ is a learning pace controlling the progress of the curriculum, $s$ is the training step, and $\lambda_0$ is the initial weight for integrity.

After selecting these responses, we rewrite their original instructions by removing the unmet constraints to construct hindsight instruction-response buffer $\mathcal{H}$, while still retaining the original pairs in the training data buffer. Specifically, the rewritten instruction $q' = x \odot c_1 \odot \dots \odot c_j \ (c_i \in \mathcal{C}')$, where $\odot$ denotes the string concatenation operation, and $\mathcal{C}' = \{c \in \mathcal{C} \mid \mathbb{I}(q, y, c) = 1\}$ denotes the subset of original constraints \(\mathcal{C}\) that are satisfied by the response $y$. With this modification, the failed samples is assigned a non-zero reward (set to 1 in this work) and thus facilitate learning. The \textit{select}-then-\textit{rewrite} process is outlined in Algorithm~\ref{alg:selection}.

\subsection{Reinforcement Learning Objective}
In each sampling group $\mathcal{G}$, we first generate $m$ responses for an instruction and then select $k (k < m)$ failed responses for replay. If the number of failed responses $z$ is smaller than $k$, we additionally generate $k - z$ supplementary samples. Finally, the model is fine-tuned on a mixed set of both the initial and replayed samples using clear binary rewards. Our HiR training objective, adapted from the Reinforce++ algorithm~\citep{hu2025reinforce++}, is given by:
\begin{equation}
\label{eq:hir}
\begin{small}
\begin{aligned}
&\mathcal{J}_{\text{HiR}}(\theta) = \mathbb{E}_{q \sim \mathcal{D}, \{y^{(i)}\}_{i=1}^{m} \sim \pi_{\text{old}}(\cdot|q), \textcolor{red}{\{q'^{(i)}, y'^{(i)}\}_{i=1}^{k} \sim \mathcal{H}}} \\
&\bigg[ \underbrace{\frac{1}{m} \sum_{i=1}^{m} \frac{1}{|y^{(i)}|} \sum_{t=1}^{|y^{(i)}|} \min \left(\rho_{t, \theta}^{(i)} A_{t}^{(i)}, \text{clip}(\rho_{t, \theta}^{(i)}, 1 \pm \epsilon) A_{t}^{(i)} \right)}_{\text{Objective for Initial Samples}} +\\
&\underbrace{{\frac{1}{k} \sum_{i=1}^k \frac{1}{|y'^{(i)}|}\! \sum_{t=1}^{|y'^{(i)}|} \!\colorbox{myblue}{$\min \left(\rho_{t, \theta}'^{(i)} A_{t}'^{(i)}, \text{clip}(\rho_{t, \theta}'^{(i)}, 1 \pm \epsilon) A_{t}'^{(i)} \right)$}}}_{\text{Objective for Replayed Samples}} \bigg],
\end{aligned}
\end{small}
\end{equation}
where $\mathcal{D}$ is the dataset of instructions, $\mathcal{H}$ is hindsight replay buffer that contains the hindsight pseudo-instruction $q'^{(i)}$ and corresponding response $y'^{(i)}$, $A_t$ denotes the advantage term for the $t$-th token in a response that are calculated based on reward. Notably, $\rho_{t, \theta}^{(i)}$ and $\rho_{t, \theta}'^{(i)}$ are the token-level importance sampling ratio between the current policy $\pi_\theta$ and old policy $\pi_{\text{old}}$:
\begin{equation}
\begin{aligned}
\rho_{t, \theta}^{(i)} = \frac{\pi_\theta(y_{t}^{(i)} | q, y_{<t}^{(i)})}{\pi_{\text{old}}(y_{t}^{(i)} | q, y_{<t}^{(i)})}, \
\rho_{t, \theta}'^{(i)} = \frac{\pi_\theta(y_{t}^{(i')} | \textcolor{red}{q'^{(i)}}, y_{<t}'^{(i)})}{\pi_{\text{old}}(y_{t}^{(i)} | q, y_{<t}^{(i)})}.
\end{aligned}
\end{equation}
Algorithm~\ref{alg:hir} presents the complete HiR training procedure.

\begin{algorithm}[!h]
\caption{Hindsight Instruction Replay}
\label{alg:hir}
\begin{algorithmic}[1]
\REQUIRE Initial policy $\pi_\theta$, Training batch data $\mathcal{D}$
\STATE \textbf{Input:} $m$, $k$, $\eta$, $\lambda_0$, reward function $r(\cdot)$
\STATE Initialize $\pi_{ref} \leftarrow \pi_\theta$, $\lambda \leftarrow \lambda_0$
\FOR{each training step $s$}
    \STATE  Experience buffer $\mathcal{B} \leftarrow \emptyset$
    \FOR{each $(q, \mathcal{C}) \sim \mathcal{D}$}
        \STATE Sampling group $\mathcal{G} \leftarrow \emptyset$
        \FOR{$i = 1$ to $m$}
            \STATE Sample response $y_i \sim \pi_\theta(\cdot|q)$
            \STATE Calculate reward $r_i \leftarrow \text{ILA}(q, y_i, \mathcal{C})$
            \STATE Store the tuple $(q, y_i, r_i)$ in buffer $\mathcal{B}, \mathcal{G}$
        \ENDFOR
        \STATE \textcolor{OliveGreen}{\textit{// Select a subset $\mathcal{T}$ of $\mathcal{G}$ for replay by Alg. 1}}
        \STATE $\mathcal{H} \leftarrow$ \textsc{Select-Rewrite} $(\mathcal{G}, k)$
        \FOR{each tuple ($q_k$, $y_k$, $\mathcal{C}_k$) in $\mathcal{H}$}
            \STATE \textcolor{OliveGreen}{\textit{// Replay the response under pseudo-instruction}}
            \STATE Calculate reward $r_k \leftarrow \text{ILA}(q_k, y_k, \mathcal{C}_k)$
            \STATE Store the tuple $(q_k, y_k, r_k)$ in buffer $\mathcal{B}$
        \ENDFOR
    \ENDFOR
    \STATE Compute advantages $A_i$ based on rewards
    \STATE Update policy model $ \pi_\theta$ using experience buffer $\mathcal{B}$
    \STATE Update $\lambda \leftarrow (1+\eta)^s \cdot \lambda_0$
\ENDFOR
\STATE \textbf{Return:} Trained policy $\pi_\theta$
\end{algorithmic}
\end{algorithm}

\subsection{Theoretical Perspective}
In this section, we re-examine the training objective of HiR from the perspective of preference learning. This perspective clarifies the underlying mechanism of HiR: it not only learns preference on different responses but also motivates a deeper investigation into the preference of instructions relative to a response. We first formulate the preference-based objective inspired by~\citet{rafailov2023direct}.
\begin{definition}
Let $\pi_\theta$ be the language model, and $\mathcal{X}, \mathcal{Y}$ be the input and output distribution, respectively. We define the positive sample $\bf{y^+} \in \mathcal{Y}$ and negative sample $\bf{y^-} \in \mathcal{Y}$ when $\pi_\theta$ receives a prompt $\bf{x} \in \mathcal{X}$, where $\bf{y^+}$ is preferred to $\bf{y^-}$ according to an underlying reward function. We define the preference learning objective as:
\begin{equation}
    \mathcal{J}(\theta) = - \mathbb{E}_{\bf{x}, \bf{y^+}, \bf{y^-}} \ [\alpha \cdot \pi_\theta({\bf{y^+} \mid \bf{x}}) - \beta \cdot \pi_\theta(\bf{y^-} \mid \bf{x})].
\end{equation}
Here $\alpha$ and $\beta$ are both positive coefficients that weight positive and negative contributions.
\end{definition}
\begin{proposition}
\label{pro:constract}
The HiR objective is a form of preference learning on both the response- and instruction-level.
\begin{equation}
\begin{aligned}
& J_{\text{HiR}}(\theta) = {\mathbb{E}}_{q \sim \mathcal{D}, q' \sim \mathcal{H}} \biggl[ \\
& \biggr( \underbrace{\alpha_1 \underset{y^w \sim \pi_{\theta}(\cdot \mid q)}{\mathbb{E}} \pi_\theta(\textcolor{OliveGreen}{y^w} \mid q) - \beta_1 \underset{y^l \sim \pi_{\theta}(\cdot \mid q)}{\mathbb{E}} \pi_\theta(\textcolor{red}{y^l} \mid q)}_{\text{Response-level Preference}}  \biggr) + \\ &\biggr( \underbrace{\alpha_2 \underset{y^r \sim \pi_{\theta}(\cdot \mid q)}{\mathbb{E}} \pi_\theta(y^r \mid \textcolor{OliveGreen}{q'}) - \beta_2 \underset{y^r \sim \pi_{\theta}(\cdot \mid q)}{\mathbb{E}} \pi_\theta(y^r \mid \textcolor{red}{q})}_{\text{Instruction-level Preference}}  \biggr) \biggr]
\end{aligned}
\end{equation}
where $y^w$ and $y^l$ denote the winning (positive) and losing (negative) responses, $y^r$ denotes the responses that are selected for replay, $\alpha_1, \alpha_2, \beta_1, \beta_2$ are all positive values calculated based on the rewards of samples.
\end{proposition}
\textbf{Remark}. Proposition~\ref{pro:constract} establishes a unified view of HiR as a dual-preference learning. While the first term aligns with standard preference on winning and losing responses, the second term introduces a discriminative signal in the instruction space. By contrasting the preference of a response under the hindsight pseudo-instruction $q'$ against the original instruction $q$, the model is encouraged to capture subtle distinctions between instructions. The detailed proof can be found in Appendix~\ref{app:proof}.

\section{Experiments}
\subsection{Experimental Setup}
\textbf{Datasets and Benchmark.} The training dataset aims to improve the capabilities of LLMs in complex instruction following tasks, while balancing quantity, diversity, and quality. To this end, we collect public data from various sources, including MulDimIF~\citep{ye2025multi}, VerIF~\citep{peng2025verif}, IFTrain~\citep{Pyatkin2025GeneralizingVI}, and Chatbot Arena~\citep{zheng2023judging}. We further synthesize constraints using programmatic approaches to enrich the dataset. After selection and construction, we obtained the \textbf{\texttt{HIR-16K}} dataset, which consists of 16K queries in different scenarios, each paired with more than 5 decomposable constraints. We employ seven public benchmarks to evaluate the instruction following alibity, including IFEval~\citep{zhou2023instruction}, IFBench~\citep{Pyatkin2025GeneralizingVI}, CFBench~\citep{zhang2025cfbench}, InfoBench~\citep{qin2024infobench}, ComplexBench~\citep{wen2024benchmarking}, MulDimIF~\citep{ye2025multi} and FollowBench~\citep{jiang2024followbench}. Additionally, we test on three out-of-domain popular reasoning benchmarks to measure its general capabilities: MATH-500~\citep{lightman2023let}, GPQA~\citep{rein2024gpqa} and MMLU-Pro~\citep{mmlupro2024wang}. Detailed dataset information is presented in Appendix~\ref{app:data_info}.

\textbf{Models and Configurations.}
We choose multiple initial models of different backbones and parameter scales for our experiments, including the Qwen2.5 series~\citep{yang2025qwen25} (Qwen2.5-7B-Instruct), Qwen3 series~\citep{yang2025qwen3} (Qwen3-4B-Instruct-2507), and Llama3.2 series~\citep{meta2024llama} (Llama3.2-3B-Instruct). We use verl framework~\citep{sheng2025hybridflow} to conduct RL training experiments on both baselines and our algorithm. For the implementation of replay strategy, we set $\eta = 0.05$, $\lambda_0 = 2$, $m = 6$, and $k=2$ in Algorithm~\ref{alg:hir}. We use DeepSeek-V3.1 (non-thinking)~\citep{liu2024deepseek} as the judge LLM for both training and evaluation. More detailed training and evaluation hyperparameters can be found in Appendix~\ref{app:implementation}.

\textbf{Baselines and Evaluation Metrics.} We compare HiR against three categories of baselines in our experiments: (1) \textit{SFT}: Supervised Fine Tuning on GPT-5 generated responses of our training data; (2) \textit{DPO}: Direct Preference Optimization~\citep{rafailov2023direct} on pairs of chosen and rejected responses generated by GPT-5 and Qwen2.5-7B-Instruct, respectively; (3) \textit{RL}: Reinforcement Learning with instruction-level accuracy as reward (\textit{RL-IR})~\citep{peng2025verif} and constraint-level accuracy as reward (\textit{RL-CR})~\citep{qi2025constraint, Pyatkin2025GeneralizingVI} on our training data. We evaluate the performance of each model by reporting its instruction-level accuracy (Eq.~\ref{eq:pla}), which is the percentage of prompts that satisfy all given constraints. 

\begin{table*}[t!]
\centering
\small
\caption{Results on diverse instruction following dataset with different LLMs. \underline{Underline} represents the best performance among all baselines, \textbf{bold} represents the best performance among all methods, and arrow indicates \textcolor{OliveGreen}{improvement} or \textcolor{Maroon}{degradation} over the initial model, and $\dagger$ denotes the best performance among frontier models.}
\resizebox{1 \linewidth}{!}{
\begin{tabular}{l|ccccccc}
\toprule
Model & \textbf{IFEval} & \textbf{IFBench} & \textbf{CFBench} & \textbf{InfoBench} & \textbf{ComplexBench} & \textbf{MulDimIF} & \textbf{FollowBench} \\
\midrule
\rowcolor{gray!8}
\multicolumn{8}{c}{\textit{Frontier Models}} \\
\midrule
GPT-4.1 & 87.8 & 39.5 $^\dagger$ & 73.2 & 60.6 $^\dagger$ & 65.7 & 70.3 $^\dagger$ & 86.0 $^\dagger$ \\
DeepSeek-V3.1 & 86.1 & 34.7 & 75.6 $^\dagger$ & 58.4 & 66.8 $^\dagger$ & 68.3 & 83.5 \\
Gemini-2.5-Flash & 89.3 $^\dagger$ & 36.1 & 72.8 & 57.4 & 64.4 & 70.1 & 78.5 \\
\midrule
\rowcolor{gray!8}
\multicolumn{8}{c}{\textit{Our Models}} \\
\midrule
Llama-3.2-3B-Instruct & 71.2 \basex{0.0} & 23.8 \basex{0.0} & 31.3 \basex{0.0} & 44.8 \basex{0.0} & 27.6 \basex{0.0} & 35.8 \basex{0.0} & 58.0 \basex{0.0} \\
\hspace*{1em}+ SFT & 73.0 \uprise{1.8} & 24.8 \uprise{1.0} & 34.6 \uprise{3.3} & \underline{47.0} \uprise{2.2} & 26.4 \down{1.2} & 66.9 \uprise{31.1} & 58.1 \uprise{0.1} \\
\hspace*{1em}+ DPO &  74.3 \uprise{3.1} & 22.1 \down{1.7} & \underline{40.1} \uprise{8.8} & 44.4 \down{0.4} & 31.2 \uprise{3.6} & 54.4 \uprise{18.6} & \underline{61.8} \uprise{3.8} \\
\hspace*{1em}+ RL-IR  & 77.6 \uprise{6.4} &  25.3 \uprise{1.5} & 39.2 \uprise{7.9} & 46.6 \uprise{1.8} & 29.8 \uprise{2.2} & 76.3 \uprise{40.5} & 60.4 \uprise{2.4}  \\
\hspace*{1em}+ RL-CR  & \underline{79.1} \uprise{7.9} & \underline{26.6} \uprise{2.8} & 38.9 \uprise{7.6} & 46.2 \uprise{1.4} & \underline{30.2} \uprise{2.6} & \underline{77.6} \uprise{41.8} & 61.1 \uprise{3.1} \\
\rowcolor{cyan!10}
\textbf{\hspace*{1em}+ HiR (Ours)} & \textbf{83.6} \uprise{\textbf{12.4}} &  \textbf{30.4} \uprise{\textbf{6.6}} & \textbf{41.8} \uprise{\textbf{10.5}} & \textbf{49.2} \uprise{\textbf{4.4}} & \textbf{31.7} \uprise{\textbf{4.1}} & \textbf{84.9} \uprise{\textbf{49.1}} & \textbf{63.6} \uprise{\textbf{5.6}} \\
\midrule
Qwen2.5-7B-Instruct & 72.6 \basex{0.0} & 26.2 \basex{0.0} & 57.5 \basex{0.0} & 49.4 \basex{0.0} & 49.1 \basex{0.0} & 51.4 \basex{0.0} & 61.5 \basex{0.0} \\
\hspace*{1em}+ SFT & 75.6 \uprise{3.0} & 27.9 \uprise{1.7} & 53.1 \down{4.4} & 48.2 \down{1.2} & 47.3 \down{1.8} & 67.8 \uprise{16.4} & 62.6 \uprise{1.1}  \\
\hspace*{1em}+ DPO & 66.9 \down{5.7} & 25.9 \down{0.3} & 58.4 \uprise{0.9} & 50.6 \uprise{1.2} & 48.9 \down{0.2} & 56.5 \uprise{5.1} & \underline{\textbf{66.7}} \uprise{\textbf{5.2}} \\
\hspace*{1em}+ RL-IR  & 76.2 \uprise{3.6} & 31.1 \uprise{4.9} & 60.1 \uprise{2.6} & 49.8 \uprise{0.4} & \underline{50.9} \uprise{1.8} & 72.2 \uprise{20.8} & 62.3 \uprise{0.8} \\
\hspace*{1em}+ RL-CR & \underline{77.3} \uprise{4.7} & \underline{31.6} \uprise{5.4} & \underline{60.8} \uprise{3.3} & \underline{51.2} \uprise{1.8} & 50.3 \uprise{1.2} & \underline{73.5} \uprise{22.1} & 63.4 \uprise{1.9} \\
\rowcolor{cyan!10}
\textbf{\hspace*{1em}+ HiR (Ours)}  & \textbf{81.0} \uprise{\textbf{8.4}} & \textbf{35.8} \uprise{\textbf{9.6}} & \textbf{64.2} \uprise{\textbf{6.7}} & \textbf{54.6} \uprise{\textbf{5.2}} & \textbf{53.3} \uprise{\textbf{4.2}} & \textbf{79.4} \uprise{\textbf{28.0}} & 65.1 \uprise{3.6} \\
\midrule
Qwen3-4B-Instruct-2507 & 83.4 \basex{0.0} & 29.9 \basex{0.0} & 67.5 \basex{0.0} & 56.8 \basex{0.0} & 57.7 \basex{0.0} & 57.3 \basex{0.0} & 76.1 \basex{0.0} \\
\hspace*{1em}+ SFT & 83.4 \basex{0.0} & 31.3 \uprise{1.4} & 64.2 \down{3.3} & 55.0 \down{1.8} & 55.9 \down{1.8} & 66.8 \uprise{9.5} & 74.9 \down{1.2} \\
\hspace*{1em}+ DPO & 83.9 \uprise{0.5} & 27.9 \down{2.0} & 68.0 \uprise{0.5} & 57.4 \uprise{0.6} & 58.1 \uprise{0.4} & 61.5 \uprise{4.2} & 78.0 \uprise{1.9} \\
\hspace*{1em}+ RL-IR & 85.0 \uprise{1.5} & 34.1 \uprise{4.2} & \underline{69.8} \uprise{2.3} & 58.0 \uprise{1.2} & 58.2 \uprise{0.5} & 78.3 \uprise{21.0} & 77.6 \uprise{1.5} \\
\hspace*{1em}+ RL-CR & \underline{85.8} \uprise{2.4} & \underline{36.9} \uprise{7.0} & 68.5 \uprise{1.0} & \underline{58.4} \uprise{1.6} & \underline{59.6} \uprise{1.9} & \underline{79.0} \uprise{21.7} & \underline{78.2} \uprise{2.1} \\
\rowcolor{cyan!10}
\textbf{\hspace*{1em}+ HiR (Ours)} & \textbf{86.3} \uprise{\textbf{2.9}} & \textbf{40.5} \uprise{\textbf{10.6}} & \textbf{73.2} \uprise{\textbf{5.7}} & \textbf{60.8} \uprise{\textbf{4.0}} & \textbf{61.5} \uprise{\textbf{3.8}} & \textbf{80.6} \uprise{\textbf{23.3}} & \textbf{80.4} \uprise{\textbf{4.3}} \\
\bottomrule
\end{tabular}}
\label{tab:main_results}
\end{table*}

\subsection{Main Results}

\textbf{HiR applies to different model backbones and achieves consistent gains.}
We conduct a comprehensive evaluation on seven instruction following benchmarks between our method and state-of-the-art baselines. As shown in Table~\ref{tab:main_results}, HiR achieves substantial improvements across different model backbones and scales, with Qwen3-4B-Instruct-2507 surpassing many leading LLMs (\textit{e.g.}, Deepseek-V3.1, GPT-4.1) on multiple benchmarks. Under the RL framework, HiR delivers the best performance on most instruction following tasks, achieving greater gains than RL with constraint-level rewards (RL-CR). Moreover, our method exhibits superior robustness and generalization ability without observed performance degradation compared to SFT and DPO. Notably, HiR is particularly effective and yields larger improvements for initially weaker models, like Llama-3.2-3B-Instruct. We attribute this advantage to our hindsight replay mechanism that converts failure responses into successful ones, thus providing more informative learning signals. As the capability of the initial model increases, performance gains on saturated metrics (\textit{e.g.}, Qwen3-4B-Instruct-2507 on IFEval) diminish, yet advantages remain pronounced on more challenging datasets, such as IFBench and MultiDimIF.

\textbf{HiR preserves general reasoning abilities in out-of-domain scenarios.}
To assess whether optimizing for instruction following capability compromises broad problem-solving competence, we evaluate our method on three out-of-domain (OOD) reasoning benchmarks that are orthogonal to instruction following. As shown in Table~\ref{tab:generalization}, although HiR is trained solely on instruction following data, it preserves the models’ OOD performances. Across all tested backbones, our method maintains parity with the initial models on these comprehensive benchmarks, with no significant drop and occasional marginal gains that fall within typical variance. These results reflect the robustness of our training data and indicate that HiR regularizes the policy toward better intent grounding and constraint satisfaction without collapsing general reasoning ability.
\begin{table}[h!]
\centering
\small
\caption{Performance of HiR on out-of-domain benchmarks.}
\resizebox{1 \linewidth}{!}{
\begin{tabular}{l|lll}
\toprule
Model & \textbf{MATH-500} & \textbf{GPQA} & \textbf{MMLU-Pro} \\
\midrule
Llama-3.2-3B-Instruct & 47.8 & 30.8 & 34.9 \\
\rowcolor{cyan!10}
\textbf{\hspace*{1em}+ HiR (Ours)} & 49.0 \uprise{1.2} & 29.5 \down{1.3} & 37.7 \uprise{2.8} \\
\midrule
Qwen2.5-7B-Instruct & 76.6 & 36.4 & 56.3 \\
\rowcolor{cyan!10}
\textbf{\hspace*{1em}+ HiR (Ours)} & 76.6 \basex{0.0} & 35.9 \down{0.5} & 56.8 \uprise{0.5} \\
\midrule
Qwen3-4B-Instruct-2507 & 86.8 & 61.8 & 69.6 \\
\rowcolor{cyan!10}
\textbf{\hspace*{1em}+ HiR (Ours)} & 88.2 \uprise{1.4} & 62.1 \uprise{0.3} & 67.2 \down{2.4} \\
\bottomrule
\end{tabular}}%
\label{tab:generalization}
\end{table}

\begin{figure*}[!t]
\centering
\begin{subfigure}{0.31\textwidth}
    \centering
    \small
    \includegraphics[width=\textwidth]{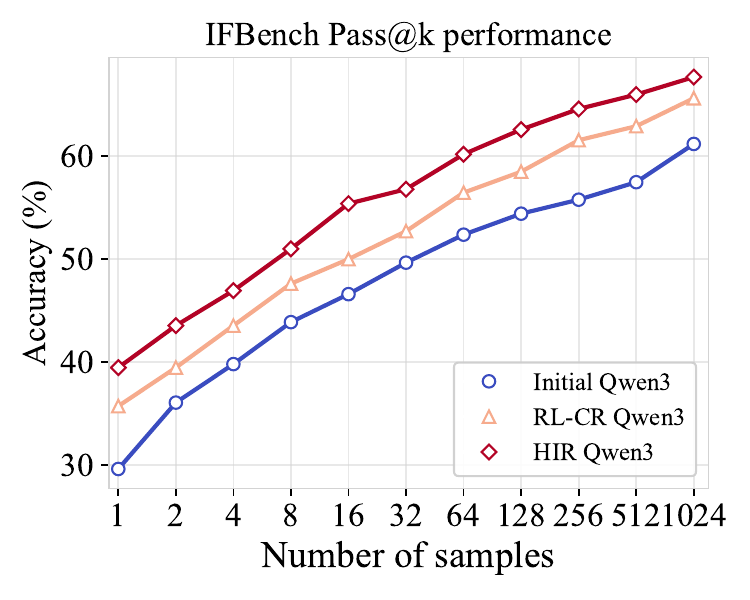}
    \caption{Qwen3-4B-Instruct-2507.}
    \label{fig:passk}
\end{subfigure}
\hfill
\begin{subfigure}{0.67\textwidth}
    \centering
    \small
    \includegraphics[width=\textwidth]{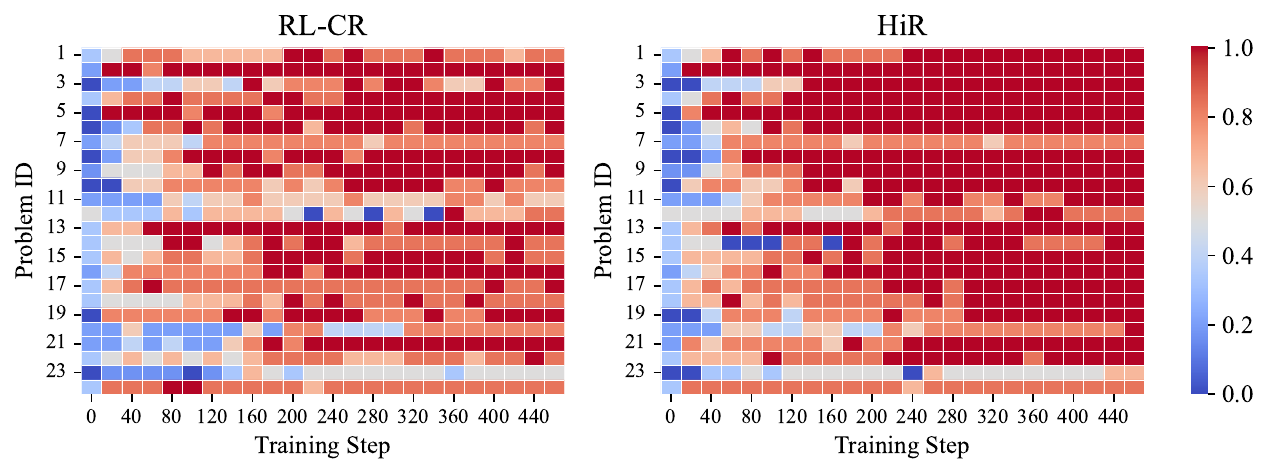}
    \caption{Llama3.2-3B-Instruct on a subset of MulDimIF.}
    \label{fig:case_heat}
\end{subfigure}
\caption{(a) The pass@$k$ curves comparison after training, and (b) constraint-level accuracy heatmap comparison during training.}
\label{fig:ablation_studies}
\end{figure*}

\begin{table*}[t!]
\centering
\small
\caption{Ablation study of selection strategy. \textbf{Bold} represents the best performance among all methods.}
\resizebox{1 \linewidth}{!}{
\begin{tabular}{l|ccccccc}
\toprule
Method & \textbf{IFEval} & \textbf{IFBench} & \textbf{CFBench} & \textbf{InfoBench} & \textbf{ComplexBench} & \textbf{MulDimIF} & \textbf{FollowBench} \\

\midrule
\rowcolor{gray!8}
\multicolumn{8}{c}{Model I: Llama-3.2-3B-Instruct} \\
\midrule

\textit{w/} Random replay
& 79.9 & 28.2 & 40.1 & 47.8 & 30.9 & 83.3 & 62.4 \\
\textit{w/} \textbf{HiR (Ours)} & \textbf{83.6} & \textbf{30.4} & \textbf{41.8} & \textbf{49.2} & \textbf{31.7} & \textbf{84.9} & \textbf{63.6} \\

\midrule
\rowcolor{gray!8}
\multicolumn{8}{c}{Model II: Qwen2.5-7B-Instruct} \\
\midrule

\textit{w/} Random replay & 79.5 & 33.7 & 63.3 & 53.6 & 51.7 & 78.1 & \textbf{66.2} \\
\textit{w/} \textbf{HiR (Ours)} & \textbf{81.0} & \textbf{35.8} & \textbf{64.2} & \textbf{54.6} & \textbf{53.3} & \textbf{79.4} & 65.1 \\

\midrule
\rowcolor{gray!8}
\multicolumn{8}{c}{Model III: Qwen3-4B-Instruct-2507} \\
\midrule

\textit{w/} Random replay & 85.2 & 38.8 & 72.5 & 59.6 & 60.9 & 79.8 & 79.5 \\
\textit{w/} \textbf{HiR (Ours)} & \textbf{86.3} & \textbf{40.5} & \textbf{73.2} & \textbf{60.8} & \textbf{61.5} & \textbf{80.6} & \textbf{80.4} \\
\bottomrule
\end{tabular}}
\vspace{-5pt}
\label{tab:ablation}
\end{table*}

\textbf{HiR enhances both the sampling stability and reasoning boundaries.}
Beyond Pass@$1$ scores, we analyze Pass@$k$ curves to characterize the reasoning boundary under increasing sampling budgets. As shown in Figure~\ref{fig:passk}, HiR consistently outperforms the initial model and RL-CR as $k$ grows, demonstrating an expanded capability ceiling and improved sample efficiency. To better understand the learning dynamics and how its abilities evolve over time, we visualize the constraint-level accuracies on a subset of MultiDimIF across the training process for both HiR and RL-CR. The heatmap of HiR (Figure~\ref{fig:case_heat}) exhibits a smooth transition from low- to high-accuracy regions, indicating a consistent and stable improvement in instruction following capability rather than reliance on stochastic or sudden gains. Besides, we observe pronounced peaks for some problems, which suggests that HiR maintains the competence with minimal fluctuation once it masters a task. In contrast, the heatmap of RL-CR shows higher variability. While a few problems converge rapidly, others remain at fluctuating accuracy levels even after extensive training, revealing potential instability in its learning process. Overall, these analyses indicate that HiR delivers robust and consistent gains, leading to more reliable improvements while extending the achievable boundary.

\subsection{Ablation Study}
\textbf{Selection Strategy.}
We first analyze the contribution of selection strategies for hindsight instruction replay. Concretely, we adopt \textit{Random Selection} strategy which arbitrarily picks a proportion of $k/m$ samples to replay. As shown in Table~\ref{tab:ablation}, our selection strategy performs optimally in most benchmarks. This confirms that not all failed attempts are equally informative across different learning stages, and our efficiency can be attributed to a more adaptive selection of suitable samples for replay.
\begin{figure}[h]
    \centering
    \includegraphics[width=\linewidth]{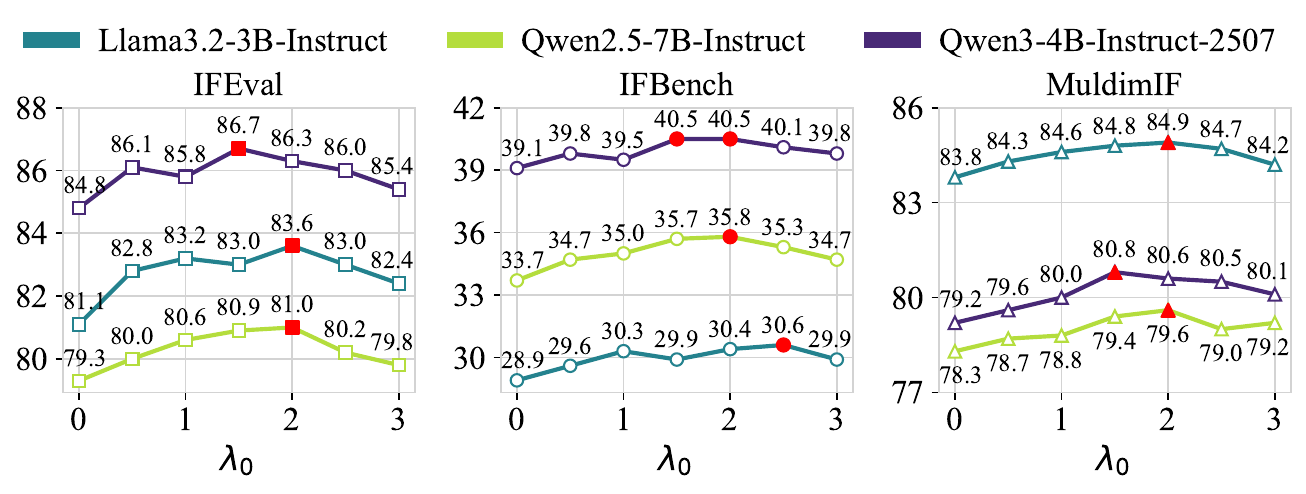}
    \vspace{-10pt}
    \caption{Ablation study of the initial curriculum weight, with \textcolor{red}{red} markers indicating the best performance for each model.}
    \vspace{-20pt}
    \label{fig:ablation}
\end{figure}

\begin{figure*}[!t]
\centering
\begin{subfigure}{\textwidth}
    \centering
    \small
    \includegraphics[width=\textwidth]{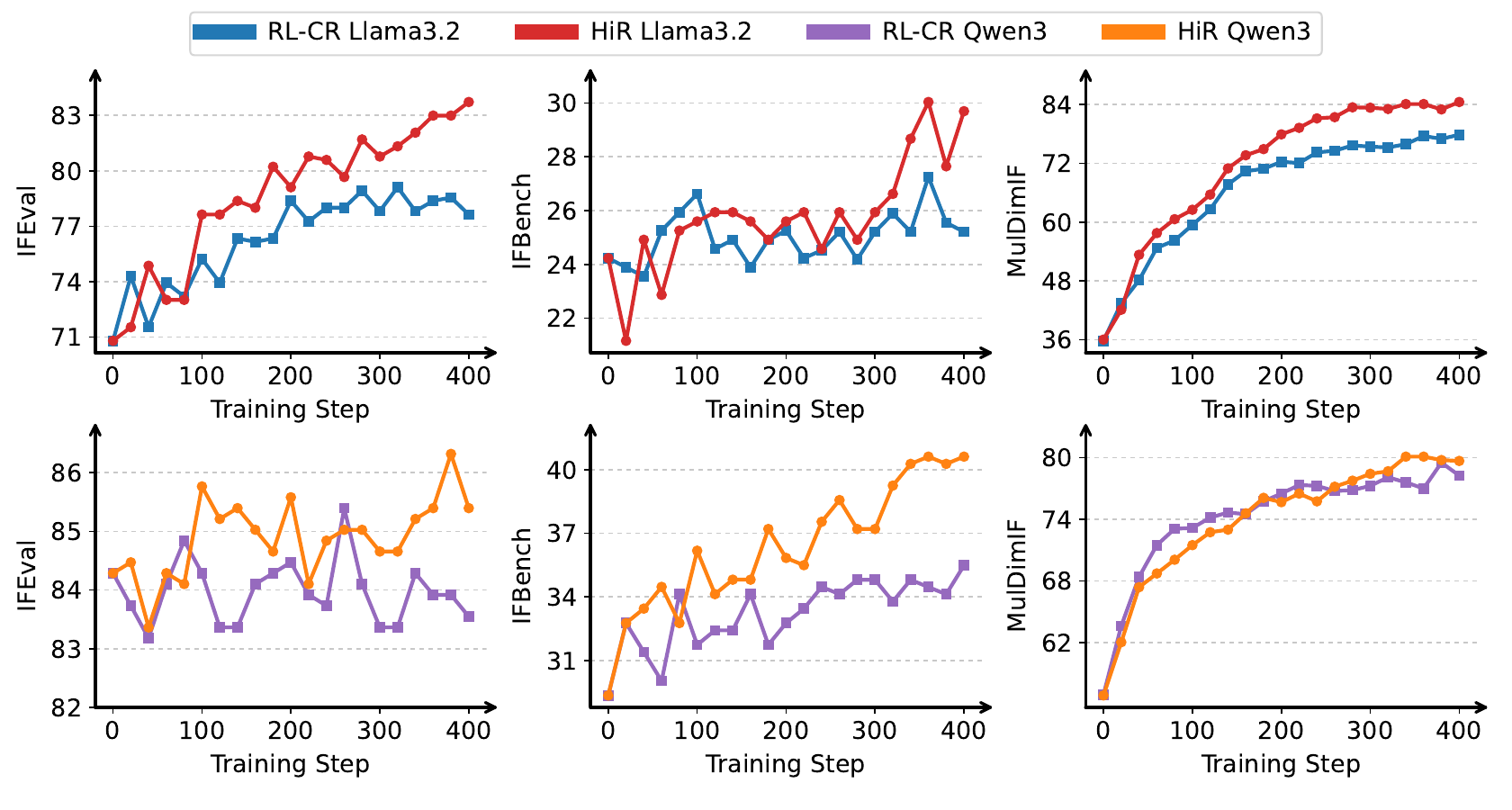}
    \label{fig:dynamics}
    \vspace{-20pt}
\end{subfigure}
\hspace*{0.002\textwidth}
\begin{subfigure}{0.33\textwidth}
    \centering
    \small
    \includegraphics[width=\textwidth]{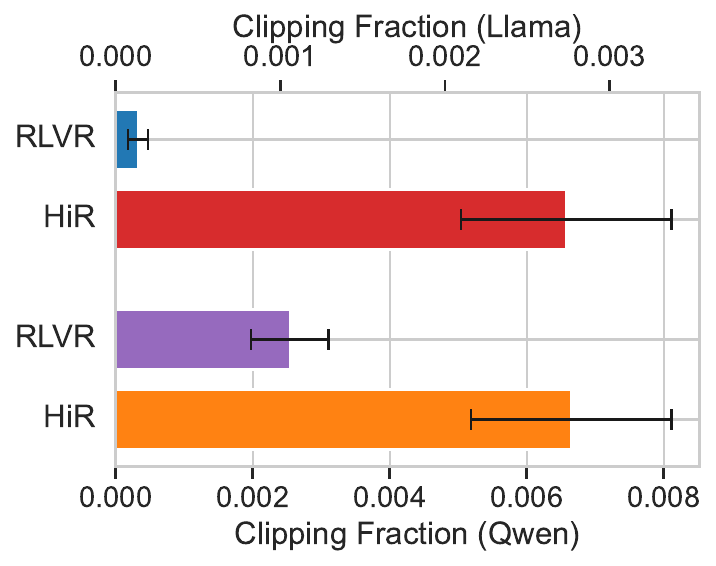}
    \label{fig:clipfrac}
\end{subfigure}
\hspace*{0.002\textwidth}
\begin{subfigure}{0.322\textwidth}
    \centering
    \small
    \includegraphics[width=\textwidth]{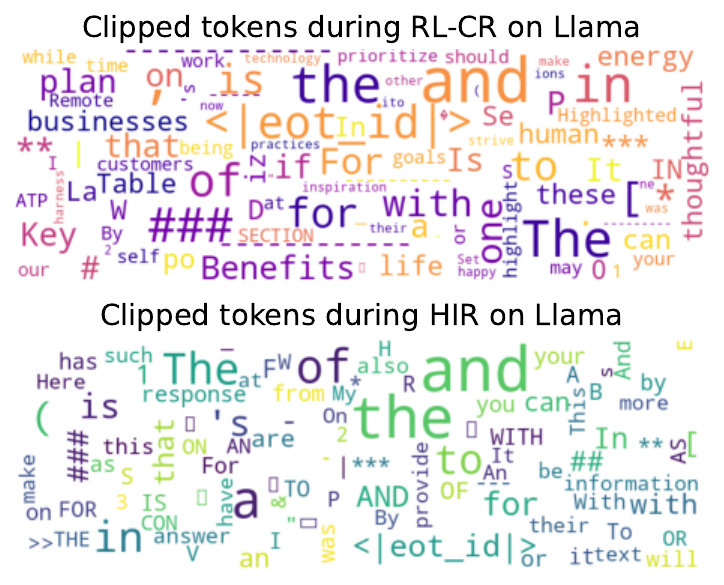}
    \label{fig:worldcloud}
\end{subfigure}
\hfill
\begin{subfigure}{0.324\textwidth}
    \centering
    \small
    \includegraphics[width=\textwidth]{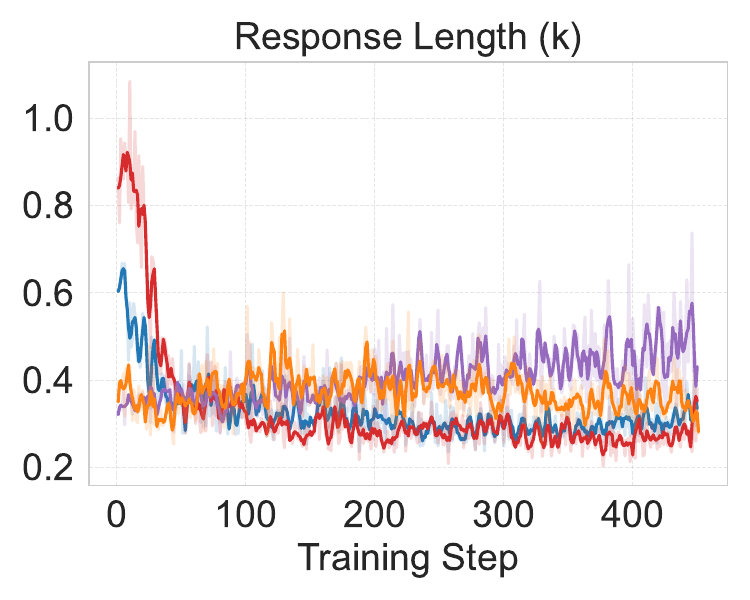}
    \label{fig:len_curve}
\end{subfigure}
\vspace{-25pt}
\caption{Training curves of different model backbones. HiR exhibits higher training efficiency than baseline RL-CR.}
\vspace{-5pt}
\label{fig:dynamic}
\end{figure*}
\textbf{Curriculum Schedule.}
To understand how the trade-off between response diversity and constraint integrity impacts final performance, we plot benchmark accuracy training with different initial curriculum weight $\lambda_0$. As shown in Figure~\ref{fig:ablation}, HiR outperforms the baseline RL-CR (in Table~\ref{tab:main_results}) over a wide range, with pronounced performance degradation only when $\lambda_0$ is excessively small or large. This phenomenon highlights a trade-off between exploration and exploitation: emphasizing constraint integrity too early (large $\lambda_0$) may lead to insufficient exploration of the solution space; while prioritizing response diversity overlong (small $\lambda_0$) may fail to provide the necessary guidance required to satisfy all constraints in the later training stage. Notably, we observe that the optimal performance is stably located around $\lambda_0=2$ across various model backbones and tasks, which demonstrates that our method is robust rather than overly sensitive to hyperparameter choices.

\section{In-Depth Analysis}
\textbf{Training Dynamics.} We report the training response length curves, clipping fraction as well as the model performance curves on the different benchmarks during training. Figure~\ref{fig:dynamic} demonstrates that the training process with HiR remains stable and does not lead to longer responses, which confirms that the improvement of HiR does not come from using more tokens. Moreover, HiR also shows superior training efficiency compared to vanilla RL, achieving better benchmark performance under the same consumed prompts and fewer computational budgets. A more interesting observation is that despite clipping more tokens due to replay and therefore using fewer for training, HiR achieves higher training efficiency than vanilla RL. This finding further reveals that token-level gradient estimates may be inherently noisy and inefficient for sample exploitation. For example, the clipped tokens during RL-CR contain some key information relevant to the instructions like substantive words ``plan" and ``benefits". In contrast, HiR tends to clip the gradient of less informative transitional or connective tokens, providing a more reliable and effective learning signal.

\begin{figure}[!h]
\vspace{-10pt}
\centering
\begin{subfigure}{0.242\textwidth}
    \centering
    \small
    \includegraphics[width=\textwidth]{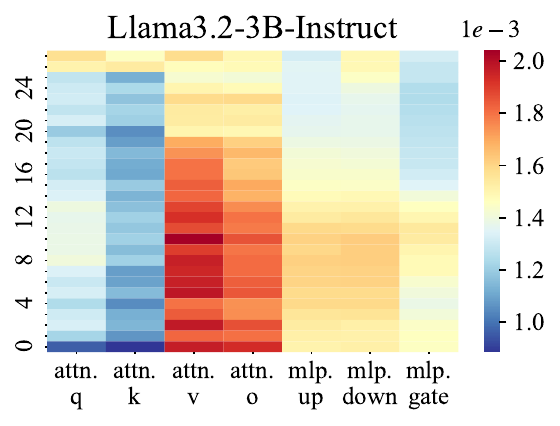}
\end{subfigure}
\hspace{-0.016\textwidth}
\begin{subfigure}{0.246\textwidth}
    \centering
    \small
    \includegraphics[width=\textwidth]{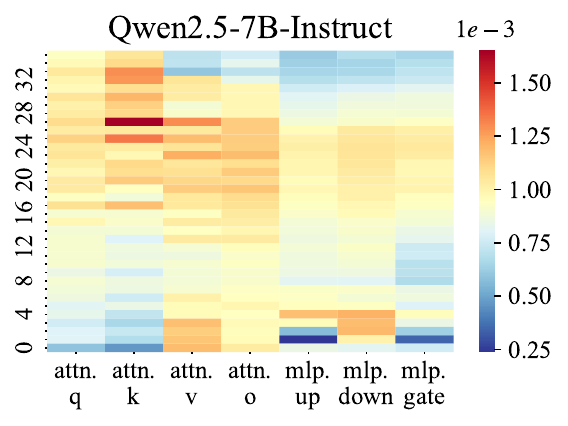}
\end{subfigure}
\caption{Parameter change over each module after HiR training.}
\vspace{-10pt}
\label{fig:parameter}
\end{figure}

\begin{table*}[t!]
\centering
\caption{A visualization of average attention allocated to each input token during generation, with \colorbox[HTML]{A9CEE4}{darker} representing greater attention.}
\vspace{-5pt}
\resizebox{\linewidth}{!}
{
\begin{tabular}{@{}p{0.33\linewidth}<{\centering}
                 p{0.33\linewidth}<{\centering}
                 p{0.33\linewidth}<{\centering}@{}}
\toprule
\textbf{Case (Initial)} & \textbf{Case (\textit{w/} RL-CR)} & \textbf{Case (\textit{w/} HiR)} \\
\midrule
\rowcolor{gray!8} \multicolumn{3}{p{\linewidth}}{\textit{\textbf{Llama-3.2-3B-Instruct:} After tuning by HiR, the model imposes greater attention to constraint `capital letters' and content of the given sentence, ensuring both format adherence and content coherence.}} \\
\begin{minipage}[h]{\linewidth}
		\centering
		{\includegraphics[width=\linewidth]{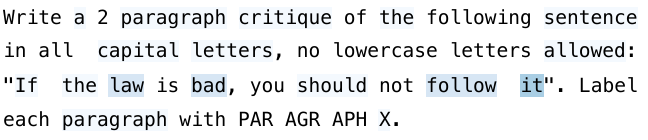}}
\end{minipage}
 & 
\begin{minipage}[h]{\linewidth}
		\centering
		{\includegraphics[width=\linewidth]{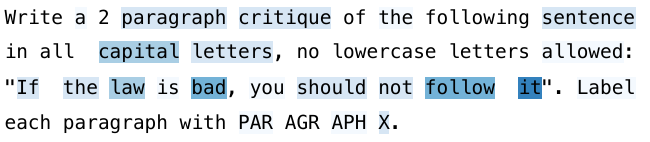}}
\end{minipage}
 & 
\begin{minipage}[h]{\linewidth}
		\centering
		{\includegraphics[width=\linewidth]{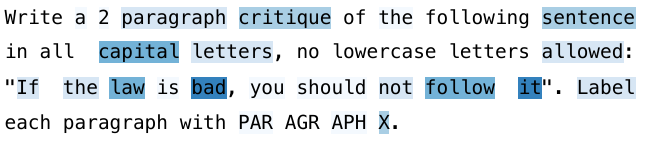}}
\end{minipage}
\\ 
\midrule
\rowcolor{gray!8} \multicolumn{3}{p{\linewidth}}{\textit{\textbf{Qwen3-4B-Instruct-2507:} After tuning by HiR, the model places more emphasis on keywords `compensated' and `immigrants' while reducing the attention to less informative pronoun `me', enabling to concentrate on the key information.}} \\
\begin{minipage}[h]{\linewidth}
		\centering
		{\includegraphics[width=\linewidth]{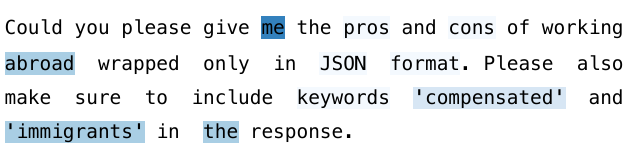}}
\end{minipage}
 & 
\begin{minipage}[h]{\linewidth}
		\centering
		{\includegraphics[width=\linewidth]{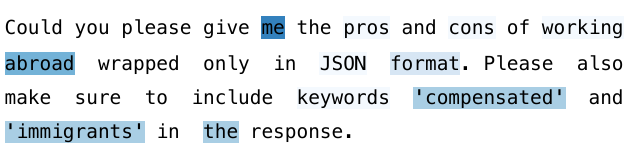}}
\end{minipage}
 & 
\begin{minipage}[h]{\linewidth}
		\centering
		{\includegraphics[width=\linewidth]{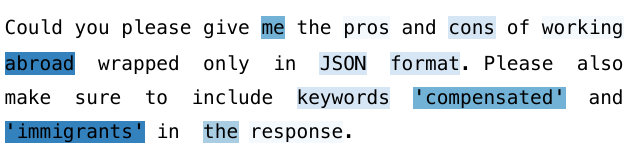}}
\end{minipage}
\\ 
\bottomrule
\end{tabular}
}
\label{tab:case}
\vspace{-10pt}
\end{table*}

\textbf{Parameter Change.}
To investigate the underlying sources of performance gain, we conduct a parameter-level analysis following~\citet{ye2025multi}. We quantified the relative change rate $|W_{\text{Init}} - W_{\text{HiR}}| / |W_{\text{Init}}|$ in model parameters after HiR training, and group the values by different modules. As depicted in Figure~\ref{fig:parameter}, we observe that most significant updates occurred within the value modules of self-attention. This suggests that HiR primarily optimizes how the model ``attends" to given information. Moreover, these variations were uniformly distributed across all layers, indicating a global rather than local adjustment. Therefore, the improvement of HiR may stem from an enhanced capacity to identify and exploit critical input tokens during training, thereby boosting its instruction following performance.

\textbf{Attention Attribution.}
To provide deeper insights into the evolution of attention mechanisms, we compute the average attention allocated to each input token during generation. As shown in Table~\ref{tab:case}, HiR training drives a more pronounced increase in attention toward constraint-related tokens than RL-CR, while simultaneously diminishing attention toward irrelevant tokens. This suggests that the model has a refined discriminative capability to identify critical constraint information while suppressing noise from distracting elements. These qualitative results empirically validate the superiority of our HiR framework, guiding the model toward more robust performance gains compared to vanilla RL approaches.

\section{Related Work}
\textbf{Instruction Following Methods.}
Early approaches primarily focused on synthesizing high-quality data for instruction tuning. Complex instructions are typically generated via instruction-evolving~\citep{xu2024wizardlm,dongself} or back-translation from existing corpora~\citep{liu2025air,qi2025constraint}. Subsequently, rejection sampling with rules~\citep{dongself} or LLMs~\citep{chengspar,liu2025air,HuangRPO,Zhang2024IOPOEL}, is applied to curate high-quality responses or preference pairs. While effective, fine-tuning with off-the-shelf data struggles to generalize to complex, unseen instructions. Recently, Tulu3~\citep{lambert2024tulu} and subsequent works~\citep{qin2025incentivizing,peng2025verif,liu2025recast} explore reinforcement learning with verifiable rewards through ``LLM-as-a-Judge'' paradigm~\citep{li2025generation} for more generalizable instruction following. However, these approaches struggle with sampling inefficiencies and ambiguous rewards. We alleviate these issues by rewriting imperfect samples into valuable training experiences, thereby improving both sample efficiency and reward clarity.

\textbf{Hindsight Experience Replay.}
Hindsight Experience Replay (HER) is a technique in traditional reinforcement learning designed to mitigate sparse rewards and reduce the need for complex reward engineering. \citet{andrychowicz2017hindsight} first introduces HER to replay failed experiences by replacing original goals with achieved states in the environment. On top of HER, several subsequent works have been proposed to encourage better exploration in environment~\citep{fang2019curriculum, liu2019competitive}, and identify trajectories with higher energy to benefit training~\citep{zhao2018energy, nguyen2019hindsight}. DHER~\citep{fang2018dher} further extends training from static goals to complex dynamic goal settings. However, prior HER-based methods have not been explored in RL training of LLMs yet as the states in LLMs are high-dimensional and semantically coherent token sequences, lacking quantifiable representations for naive goal replacement. In this work, HiR treats atomic constraints as hindsight goals in instruction space, coupled with an adaptive replay mechanism that trades off response diversity and constraint integrity alongside the model's learning progress.

\section{Conclusion}
This work proposes HiR, a simple and efficient method to incentivize the capability of LLMs for solving complex instructions. HiR employs a \textit{select}-then-\textit{rewrite} strategy that adaptively selects failure samples in a curriculum-based manner, followed by rewriting their instructions into ``hindsight" pseudo-instructions for replay. In this way, HiR implicitly introduces an instruction-wise preference into the RL training objective, enabling LLMs to precisely identify unmet constraints in instructions for effective learning with only a binary reward. Extensive experiments demonstrate that HiR consistently outperforms current baselines and achieves competitive results compared with leading models. Currently we apply hindsight instruction replay to RL for LLMs, we expect to explore applications to multi-modal tasks and agentic scenarios for future work.



\bibliography{main}

\begin{thebibliography}{63}
\providecommand{\natexlab}[1]{#1}
\providecommand{\url}[1]{\texttt{#1}}
\expandafter\ifx\csname urlstyle\endcsname\relax
  \providecommand{\doi}[1]{doi: #1}\else
  \providecommand{\doi}{doi: \begingroup \urlstyle{rm}\Url}\fi

\bibitem[Andrychowicz et~al.(2017)Andrychowicz, Wolski, Ray, Schneider, Fong, Welinder, McGrew, Tobin, Pieter~Abbeel, and Zaremba]{andrychowicz2017hindsight}
Andrychowicz, M., Wolski, F., Ray, A., Schneider, J., Fong, R., Welinder, P., McGrew, B., Tobin, J., Pieter~Abbeel, O., and Zaremba, W.
\newblock Hindsight experience replay.
\newblock \emph{Advances in Neural Information Processing Systems}, 2017.

\bibitem[Arun et~al.(2023)Arun, Dhiman, Soni, and Hu]{arun2023numerical}
Arun, A., Dhiman, A., Soni, M., and Hu, Y.
\newblock Numerical reasoning for financial reports.
\newblock \emph{arXiv preprint arXiv:2312.14870}, 2023.

\bibitem[Cheng et~al.(2025)Cheng, Liu, Wang, Gu, Lu, Zhang, Dong, Tang, Wang, and Huang]{chengspar}
Cheng, J., Liu, X., Wang, C., Gu, X., Lu, Y., Zhang, D., Dong, Y., Tang, J., Wang, H., and Huang, M.
\newblock Spar: Self-play with tree-search refinement to improve instruction-following in large language models.
\newblock In \emph{The Thirteenth International Conference on Learning Representations}, 2025.

\bibitem[Dong et~al.(2025)Dong, Lu, Li, Xia, Yu, Zhou, and Zhou]{dongself}
Dong, G., Lu, K., Li, C., Xia, T., Yu, B., Zhou, C., and Zhou, J.
\newblock Self-play with execution feedback: Improving instruction-following capabilities of large language models.
\newblock In \emph{The Thirteenth International Conference on Learning Representations}, 2025.

\bibitem[Driess et~al.(2023)Driess, Xia, Sajjadi, Lynch, Chowdhery, Ichter, Wahid, Tompson, Vuong, Yu, Huang, Chebotar, Sermanet, Duckworth, Levine, Vanhoucke, Hausman, Toussaint, Greff, Zeng, Mordatch, and Florence]{krause2023palm}
Driess, D., Xia, F., Sajjadi, M. S.~M., Lynch, C., Chowdhery, A., Ichter, B., Wahid, A., Tompson, J., Vuong, Q., Yu, T., Huang, W., Chebotar, Y., Sermanet, P., Duckworth, D., Levine, S., Vanhoucke, V., Hausman, K., Toussaint, M., Greff, K., Zeng, A., Mordatch, I., and Florence, P.
\newblock Palm-e: An embodied multimodal language model.
\newblock In \emph{International Conference on Machine Learning}, volume 202, pp.\  8469--8488, 2023.

\bibitem[Fang et~al.(2018)Fang, Zhou, Shi, Gong, Xu, and Zhang]{fang2018dher}
Fang, M., Zhou, C., Shi, B., Gong, B., Xu, J., and Zhang, T.
\newblock Dher: Hindsight experience replay for dynamic goals.
\newblock In \emph{International Conference on Learning Representations}, 2018.

\bibitem[Fang et~al.(2019)Fang, Zhou, Du, Han, and Zhang]{fang2019curriculum}
Fang, M., Zhou, T., Du, Y., Han, L., and Zhang, Z.
\newblock Curriculum-guided hindsight experience replay.
\newblock \emph{Advances in Neural Information Processing Systems}, 2019.

\bibitem[Firoozi et~al.(2025)Firoozi, Tucker, Tian, Majumdar, Sun, Liu, Zhu, Song, Kapoor, Hausman, et~al.]{firoozi2025foundation}
Firoozi, R., Tucker, J., Tian, S., Majumdar, A., Sun, J., Liu, W., Zhu, Y., Song, S., Kapoor, A., Hausman, K., et~al.
\newblock Foundation models in robotics: Applications, challenges, and the future.
\newblock \emph{The International Journal of Robotics Research}, 44\penalty0 (5):\penalty0 701--739, 2025.

\bibitem[Guo et~al.(2025)Guo, Yang, Zhang, Song, Zhang, Xu, Zhu, Ma, Wang, Bi, et~al.]{guo2025deepseek}
Guo, D., Yang, D., Zhang, H., Song, J., Zhang, R., Xu, R., Zhu, Q., Ma, S., Wang, P., Bi, X., et~al.
\newblock Deepseek-r1: Incentivizing reasoning capability in llms via reinforcement learning.
\newblock \emph{arXiv preprint arXiv:2501.12948}, 2025.

\bibitem[Hammoud et~al.(2025)Hammoud, Alhamoud, Hammoud, Bou-Zeid, Ghassemi, and Ghanem]{hammoud2025train}
Hammoud, H. A. A.~K., Alhamoud, K., Hammoud, A., Bou-Zeid, E., Ghassemi, M., and Ghanem, B.
\newblock Train long, think short: Curriculum learning for efficient reasoning.
\newblock \emph{arXiv preprint arXiv:2508.08940}, 2025.

\bibitem[Hu(2025)]{hu2025reinforce++}
Hu, J.
\newblock Reinforce++: A simple and efficient approach for aligning large language models.
\newblock \emph{arXiv preprint arXiv:2501.03262}, 2025.

\bibitem[Huang et~al.(2025{\natexlab{a}})Huang, Cen, Tan, Quan, Huang, and Zhang]{huang2025graphcot}
Huang, H., Cen, M., Tan, K., Quan, X., Huang, G., and Zhang, H.
\newblock Graphcot-vla: A 3d spatial-aware reasoning vision-language-action model for robotic manipulation with ambiguous instructions.
\newblock \emph{arXiv preprint arXiv:2508.07650}, 2025{\natexlab{a}}.

\bibitem[Huang et~al.(2025{\natexlab{b}})Huang, Lin, Fang, Wu, Li, Qu, Huang, and Li]{HuangRPO}
Huang, X., Lin, T., Fang, F., Wu, Y., Li, H., Qu, Y., Huang, F., and Li, Y.
\newblock Reverse preference optimization for complex instruction following.
\newblock In \emph{Findings of the Association for Computational Linguistics, {ACL} 2025}, 2025{\natexlab{b}}.

\bibitem[Jiang et~al.(2024)Jiang, Wang, Zeng, Zhong, Li, Mi, Shang, Jiang, Liu, and Wang]{jiang2024followbench}
Jiang, Y., Wang, Y., Zeng, X., Zhong, W., Li, L., Mi, F., Shang, L., Jiang, X., Liu, Q., and Wang, W.
\newblock {F}ollow{B}ench: A multi-level fine-grained constraints following benchmark for large language models.
\newblock In \emph{Annual Meeting of the Association for Computational Linguistics}, pp.\  4667--4688, 2024.

\bibitem[Kim et~al.(2024)Kim, Muhn, and Nikolaev]{kim2024financial}
Kim, A., Muhn, M., and Nikolaev, V.
\newblock Financial statement analysis with large language models.
\newblock \emph{arXiv preprint arXiv:2407.17866}, 2024.

\bibitem[Kwon et~al.(2023)Kwon, Li, Zhuang, Sheng, Zheng, Yu, Gonzalez, Zhang, and Stoica]{kwon2025efficient}
Kwon, W., Li, Z., Zhuang, S., Sheng, Y., Zheng, L., Yu, C.~H., Gonzalez, J., Zhang, H., and Stoica, I.
\newblock Efficient memory management for large language model serving with pagedattention.
\newblock In \emph{ACM Symposium on Operating Systems Principles}, 2023.

\bibitem[Lambert et~al.(2024)Lambert, Morrison, Pyatkin, Huang, Ivison, Brahman, Miranda, Liu, Dziri, Lyu, et~al.]{lambert2024tulu}
Lambert, N., Morrison, J., Pyatkin, V., Huang, S., Ivison, H., Brahman, F., Miranda, L. J.~V., Liu, A., Dziri, N., Lyu, S., et~al.
\newblock Tulu 3: Pushing frontiers in open language model post-training.
\newblock \emph{arXiv preprint arXiv:2411.15124}, 2024.

\bibitem[Lee et~al.(2023)Lee, Pujara, Sewak, White, and Jauhar]{lee2023making}
Lee, D.-H., Pujara, J., Sewak, M., White, R., and Jauhar, S.
\newblock Making large language models better data creators.
\newblock In \emph{Conference on Empirical Methods in Natural Language Processing}, pp.\  15349--15360, 2023.

\bibitem[Li et~al.(2025{\natexlab{a}})Li, Jiang, Huang, Beigi, Zhao, Tan, Bhattacharjee, Jiang, Chen, Wu, et~al.]{li2025generation}
Li, D., Jiang, B., Huang, L., Beigi, A., Zhao, C., Tan, Z., Bhattacharjee, A., Jiang, Y., Chen, C., Wu, T., et~al.
\newblock From generation to judgment: Opportunities and challenges of llm-as-a-judge.
\newblock In \emph{Conference on Empirical Methods in Natural Language Processing}, pp.\  2757--2791, 2025{\natexlab{a}}.

\bibitem[Li et~al.(2025{\natexlab{b}})Li, Zhang, Yin, Zhang, Ou, Wu, Yin, Li, Tao, Wang, et~al.]{li2025websailor}
Li, K., Zhang, Z., Yin, H., Zhang, L., Ou, L., Wu, J., Yin, W., Li, B., Tao, Z., Wang, X., et~al.
\newblock Websailor: Navigating super-human reasoning for web agent.
\newblock \emph{arXiv preprint arXiv:2507.02592}, 2025{\natexlab{b}}.

\bibitem[Lightman et~al.(2024)Lightman, Kosaraju, Burda, Edwards, Baker, Lee, Leike, Schulman, Sutskever, and Cobbe]{lightman2023let}
Lightman, H., Kosaraju, V., Burda, Y., Edwards, H., Baker, B., Lee, T., Leike, J., Schulman, J., Sutskever, I., and Cobbe, K.
\newblock Let's verify step by step.
\newblock In \emph{International Conference on Learning Representations}, 2024.

\bibitem[Lior et~al.(2025)Lior, Yehudai, Gera, and Ein-Dor]{lior2025wildifeval}
Lior, G., Yehudai, A., Gera, A., and Ein-Dor, L.
\newblock Wildifeval: Instruction following in the wild.
\newblock \emph{arXiv preprint arXiv:2503.06573}, 2025.

\bibitem[Liu et~al.(2024)Liu, Feng, Xue, Wang, Wu, Lu, Zhao, Deng, Zhang, Ruan, et~al.]{liu2024deepseek}
Liu, A., Feng, B., Xue, B., Wang, B., Wu, B., Lu, C., Zhao, C., Deng, C., Zhang, C., Ruan, C., et~al.
\newblock Deepseek-v3 technical report.
\newblock \emph{arXiv preprint arXiv:2412.19437}, 2024.

\bibitem[Liu et~al.(2019)Liu, Trott, Socher, and Xiong]{liu2019competitive}
Liu, H., Trott, A., Socher, R., and Xiong, C.
\newblock Competitive experience replay.
\newblock In \emph{International Conference on Learning Representations}, 2019.

\bibitem[Liu et~al.(2025{\natexlab{a}})Liu, Guo, Xie, Xu, Huang, Tian, Xu, Wu, Wang, Lv, et~al.]{liu2025recast}
Liu, W., Guo, Z., Xie, M., Xu, J., Huang, Z., Tian, M., Xu, J., Wu, M., Wang, X., Lv, C., et~al.
\newblock Recast: Strengthening llms' complex instruction following with constraint-verifiable data.
\newblock \emph{arXiv preprint arXiv:2505.19030}, 2025{\natexlab{a}}.

\bibitem[Liu et~al.(2025{\natexlab{b}})Liu, He, Li, Huang, Hu, Liu, Li, Su, and Zheng]{liu2025air}
Liu, W., He, Y., Li, Y., Huang, H., Hu, C., Liu, J., Li, S., Su, W., and Zheng, B.
\newblock Air: Complex instruction generation via automatic iterative refinement.
\newblock In \emph{Proceedings of the 2025 Conference on Empirical Methods in Natural Language Processing}, pp.\  31952--31974, 2025{\natexlab{b}}.

\bibitem[Meta(2024)]{meta2024llama}
Meta, A.
\newblock Llama 3.2: Revolutionizing edge ai and vision with open, customizable models.
\newblock \emph{Meta AI Blog. Retrieved December}, 2024.
\newblock URL \url{https://ai.meta.com/blog}.

\bibitem[Minaee et~al.(2024)Minaee, Mikolov, Nikzad, Chenaghlu, Socher, Amatriain, and Gao]{minaee2024large}
Minaee, S., Mikolov, T., Nikzad, N., Chenaghlu, M., Socher, R., Amatriain, X., and Gao, J.
\newblock Large language models: A survey.
\newblock \emph{arXiv preprint arXiv:2402.06196}, 2024.

\bibitem[Nguyen et~al.(2019)Nguyen, La, and Deans]{nguyen2019hindsight}
Nguyen, H., La, H.~M., and Deans, M.
\newblock Hindsight experience replay with experience ranking.
\newblock In \emph{International Conference on Development and Learning and Epigenetic Robotics}, pp.\  1--6, 2019.

\bibitem[Peng et~al.(2025)Peng, Qi, Wang, Xu, Hou, and Li]{peng2025verif}
Peng, H., Qi, Y., Wang, X., Xu, B., Hou, L., and Li, J.
\newblock Verif: Verification engineering for reinforcement learning in instruction following.
\newblock In \emph{Proceedings of the 2025 Conference on Empirical Methods in Natural Language Processing}, pp.\  30312--30327, 2025.

\bibitem[Pyatkin et~al.(2025)Pyatkin, Malik, Graf, Ivison, Huang, Dasigi, Lambert, and Hajishirzi]{Pyatkin2025GeneralizingVI}
Pyatkin, V., Malik, S., Graf, V., Ivison, H., Huang, S., Dasigi, P., Lambert, N., and Hajishirzi, H.
\newblock Generalizing verifiable instruction following.
\newblock In \emph{Advances in Neural Information Processing Systems}, 2025.

\bibitem[Qi et~al.(2025{\natexlab{a}})Qi, Peng, Wang, Xin, Liu, Xu, Hou, and Li]{qi2025agentif}
Qi, Y., Peng, H., Wang, X., Xin, A., Liu, Y., Xu, B., Hou, L., and Li, J.
\newblock Agentif: Benchmarking instruction following of large language models in agentic scenarios.
\newblock \emph{arXiv preprint arXiv:2505.16944}, 2025{\natexlab{a}}.

\bibitem[Qi et~al.(2025{\natexlab{b}})Qi, Peng, Wang, Xu, Hou, and Li]{qi2025constraint}
Qi, Y., Peng, H., Wang, X., Xu, B., Hou, L., and Li, J.
\newblock Constraint back-translation improves complex instruction following of large language models.
\newblock In \emph{Proceedings of the 34th ACM International Conference on Information and Knowledge Management}, pp.\  2388--2398, 2025{\natexlab{b}}.

\bibitem[Qian et~al.(2023)Qian, Han, Fung, Qin, Liu, and Ji]{qian2023creator}
Qian, C., Han, C., Fung, Y., Qin, Y., Liu, Z., and Ji, H.
\newblock Creator: Tool creation for disentangling abstract and concrete reasoning of large language models.
\newblock In \emph{Findings of the Association for Computational Linguistics: EMNLP}, pp.\  6922--6939, 2023.

\bibitem[Qiao et~al.(2025)Qiao, Chen, Chen, Yu, Yin, Wang, Zhang, Li, Yin, Li, Min, Liao, Jiang, Xie, Huang, and Zhou]{qiao2025webresearcher}
Qiao, Z., Chen, G., Chen, X., Yu, D., Yin, W., Wang, X., Zhang, Z., Li, B., Yin, H., Li, K., Min, R., Liao, M., Jiang, Y., Xie, P., Huang, F., and Zhou, J.
\newblock Webresearcher: Unleashing unbounded reasoning capability in long-horizon agents.
\newblock \emph{arXiv preprint arXiv:2509.13309}, 2025.

\bibitem[Qin et~al.(2024)Qin, Song, Hu, Yao, Cho, Wang, Wu, Liu, Liu, and Yu]{qin2024infobench}
Qin, Y., Song, K., Hu, Y., Yao, W., Cho, S., Wang, X., Wu, X., Liu, F., Liu, P., and Yu, D.
\newblock Infobench: Evaluating instruction following ability in large language models.
\newblock In \emph{Findings of the Association for Computational Linguistics}, pp.\  13025--13048, 2024.

\bibitem[Qin et~al.(2025)Qin, Li, Li, Xu, Shi, Lin, Cui, Li, and Sun]{qin2025incentivizing}
Qin, Y., Li, G., Li, Z., Xu, Z., Shi, Y., Lin, Z., Cui, X., Li, K., and Sun, X.
\newblock Incentivizing reasoning for advanced instruction-following of large language models.
\newblock In \emph{Annual Conference on Neural Information Processing Systems}, 2025.

\bibitem[Qwen et~al.(2025)Qwen, :, Yang, Yang, Zhang, Hui, Zheng, Yu, Li, Liu, Huang, Wei, Lin, Yang, Tu, Zhang, Yang, Yang, Zhou, Lin, Dang, Lu, Bao, Yang, Yu, Li, Xue, Zhang, Zhu, Men, Lin, Li, Tang, Xia, Ren, Ren, Fan, Su, Zhang, Wan, Liu, Cui, Zhang, and Qiu]{yang2025qwen25}
Qwen, :, Yang, A., Yang, B., Zhang, B., Hui, B., Zheng, B., Yu, B., Li, C., Liu, D., Huang, F., Wei, H., Lin, H., Yang, J., Tu, J., Zhang, J., Yang, J., Yang, J., Zhou, J., Lin, J., Dang, K., Lu, K., Bao, K., Yang, K., Yu, L., Li, M., Xue, M., Zhang, P., Zhu, Q., Men, R., Lin, R., Li, T., Tang, T., Xia, T., Ren, X., Ren, X., Fan, Y., Su, Y., Zhang, Y., Wan, Y., Liu, Y., Cui, Z., Zhang, Z., and Qiu, Z.
\newblock Qwen2.5 technical report.
\newblock \emph{arXiv preprint arXiv:2412.15115}, 2025.

\bibitem[Rafailov et~al.(2023)Rafailov, Sharma, Mitchell, Manning, Ermon, and Finn]{rafailov2023direct}
Rafailov, R., Sharma, A., Mitchell, E., Manning, C.~D., Ermon, S., and Finn, C.
\newblock Direct preference optimization: Your language model is secretly a reward model.
\newblock \emph{Advances in neural information processing systems}, 36:\penalty0 53728--53741, 2023.

\bibitem[Rein et~al.(2024)Rein, Hou, Stickland, Petty, Pang, Dirani, Michael, and Bowman]{rein2024gpqa}
Rein, D., Hou, B.~L., Stickland, A.~C., Petty, J., Pang, R.~Y., Dirani, J., Michael, J., and Bowman, S.~R.
\newblock Gpqa: A graduate-level google-proof q\&a benchmark.
\newblock In \emph{Conference on Language Modeling}, 2024.

\bibitem[Sakai et~al.(2025)Sakai, Kamigaito, and Watanabe]{sakai2025revisiting}
Sakai, Y., Kamigaito, H., and Watanabe, T.
\newblock Revisiting compositional generalization capability of large language models considering instruction following ability.
\newblock \emph{arXiv preprint arXiv:2506.15629}, 2025.

\bibitem[Sheng et~al.(2025)Sheng, Zhang, Ye, Wu, Zhang, Zhang, Peng, Lin, and Wu]{sheng2025hybridflow}
Sheng, G., Zhang, C., Ye, Z., Wu, X., Zhang, W., Zhang, R., Peng, Y., Lin, H., and Wu, C.
\newblock Hybridflow: A flexible and efficient rlhf framework.
\newblock In \emph{Proceedings of the Twentieth European Conference on Computer Systems}, pp.\  1279--1297, 2025.

\bibitem[Song et~al.(2025)Song, Gan, Shang, and Zhao]{song2025ifir}
Song, T., Gan, G., Shang, M., and Zhao, Y.
\newblock Ifir: A comprehensive benchmark for evaluating instruction-following in expert-domain information retrieval.
\newblock \emph{arXiv preprint arXiv:2503.04644}, 2025.

\bibitem[Wang et~al.(2025)Wang, Zhao, Ding, Kuang, Wang, Cao, and Cai]{wang2025ask}
Wang, J., Zhao, Y., Ding, P., Kuang, J., Wang, Z., Cao, X., and Cai, X.
\newblock Ask, fail, repeat: Meeseeks, an iterative feedback benchmark for llms' multi-turn instruction-following ability.
\newblock \emph{arXiv preprint arXiv:2504.21625}, 2025.

\bibitem[Wang et~al.(2024)Wang, Ma, Zhang, Ni, Chandra, Guo, Ren, Arulraj, He, Jiang, Li, Ku, Wang, Zhuang, Fan, Yue, and Chen]{mmlupro2024wang}
Wang, Y., Ma, X., Zhang, G., Ni, Y., Chandra, A., Guo, S., Ren, W., Arulraj, A., He, X., Jiang, Z., Li, T., Ku, M., Wang, K., Zhuang, A., Fan, R., Yue, X., and Chen, W.
\newblock Mmlu-pro: {A} more robust and challenging multi-task language understanding benchmark.
\newblock In \emph{Conference on Neural Information Processing Systems}, 2024.

\bibitem[Wen et~al.(2024)Wen, Ke, Gu, Wu, Huang, Zhou, Li, Hu, Gao, Xu, et~al.]{wen2024benchmarking}
Wen, B., Ke, P., Gu, X., Wu, L., Huang, H., Zhou, J., Li, W., Hu, B., Gao, W., Xu, J., et~al.
\newblock Benchmarking complex instruction-following with multiple constraints composition.
\newblock \emph{Advances in Neural Information Processing Systems}, 37:\penalty0 137610--137645, 2024.

\bibitem[Wu \& Choi(2025)Wu and Choi]{wu2025invisible}
Wu, F. and Choi, Y.
\newblock The invisible leash: Why rlvr may not escape its origin.
\newblock In \emph{AI for Math Workshop@ ICML}, 2025.

\bibitem[Xie et~al.(2025)Xie, Gao, Ren, Luo, Hong, Dai, Zhou, Qiu, Wu, and Luo]{xie2025logic}
Xie, T., Gao, Z., Ren, Q., Luo, H., Hong, Y., Dai, B., Zhou, J., Qiu, K., Wu, Z., and Luo, C.
\newblock Logic-rl: Unleashing llm reasoning with rule-based reinforcement learning.
\newblock \emph{arXiv preprint arXiv:2502.14768}, 2025.

\bibitem[Xu et~al.(2024)Xu, Sun, Zheng, Geng, Zhao, Feng, Tao, Lin, and Jiang]{xu2024wizardlm}
Xu, C., Sun, Q., Zheng, K., Geng, X., Zhao, P., Feng, J., Tao, C., Lin, Q., and Jiang, D.
\newblock Wizardlm: Empowering large pre-trained language models to follow complex instructions.
\newblock In \emph{The Twelfth International Conference on Learning Representations}, 2024.

\bibitem[Yang et~al.(2025)Yang, Li, Yang, Zhang, Hui, Zheng, Yu, Gao, Huang, Lv, et~al.]{yang2025qwen3}
Yang, A., Li, A., Yang, B., Zhang, B., Hui, B., Zheng, B., Yu, B., Gao, C., Huang, C., Lv, C., et~al.
\newblock Qwen3 technical report.
\newblock \emph{arXiv preprint arXiv:2505.09388}, 2025.

\bibitem[Ye et~al.(2025)Ye, Huang, Chen, Fu, Yang, Yang, Wu, Wang, Zhou, Yang, et~al.]{ye2025multi}
Ye, J., Huang, C., Chen, Z., Fu, W., Yang, C., Yang, L., Wu, Y., Wang, P., Zhou, M., Yang, X., et~al.
\newblock A multi-dimensional constraint framework for evaluating and improving instruction following in large language models.
\newblock \emph{arXiv preprint arXiv:2505.07591}, 2025.

\bibitem[Yue et~al.(2025)Yue, Chen, Lu, Zhao, Wang, Song, and Huang]{yue2025does}
Yue, Y., Chen, Z., Lu, R., Zhao, A., Wang, Z., Song, S., and Huang, G.
\newblock Does reinforcement learning really incentivize reasoning capacity in llms beyond the base model?
\newblock In \emph{Conference on Neural Information Processing Systems}, 2025.

\bibitem[Zeng et~al.(2025)Zeng, Huang, Liu, Liu, He, Ma, and He]{zeng2025simplerl}
Zeng, W., Huang, Y., Liu, Q., Liu, W., He, K., Ma, Z., and He, J.
\newblock Simplerl-zoo: Investigating and taming zero reinforcement learning for open base models in the wild.
\newblock \emph{arXiv preprint arXiv:2503.18892}, 2025.

\bibitem[Zhang et~al.(2025{\natexlab{a}})Zhang, Yao, Lai, Huang, Fang, Tao, Song, and Liu]{zhang2025reasoning}
Zhang, K., Yao, Q., Lai, B., Huang, J., Fang, W., Tao, D., Song, M., and Liu, S.
\newblock Reasoning with reinforced functional token tuning.
\newblock \emph{arXiv preprint arXiv:2502.13389}, 2025{\natexlab{a}}.

\bibitem[Zhang et~al.(2025{\natexlab{b}})Zhang, Zuo, He, Sun, Liu, Jiang, Fan, Tian, Jia, Li, et~al.]{zhang2025survey}
Zhang, K., Zuo, Y., He, B., Sun, Y., Liu, R., Jiang, C., Fan, Y., Tian, K., Jia, G., Li, P., et~al.
\newblock A survey of reinforcement learning for large reasoning models.
\newblock \emph{arXiv preprint arXiv:2509.08827}, 2025{\natexlab{b}}.

\bibitem[Zhang et~al.(2025{\natexlab{c}})Zhang, Zhu, Shen, Luo, Zhang, Liang, Yang, Lin, Qiao, Chen, Cui, Zhang, and Zhou]{zhang2025cfbench}
Zhang, T., Zhu, C., Shen, Y., Luo, W., Zhang, Y., Liang, H., Yang, F., Lin, M., Qiao, Y., Chen, W., Cui, B., Zhang, W., and Zhou, Z.
\newblock Cfbench: {A} comprehensive constraints-following benchmark for llms.
\newblock In \emph{Proceedings of the Annual Meeting of the Association for Computational Linguistics}, pp.\  32926--32944, 2025{\natexlab{c}}.

\bibitem[Zhang et~al.(2024)Zhang, Yu, Fu, Huang, and Li]{Zhang2024IOPOEL}
Zhang, X., Yu, H., Fu, C., Huang, F., and Li, Y.
\newblock Iopo: Empowering llms with complex instruction following via input-output preference optimization.
\newblock In \emph{Annual Meeting of the Association for Computational Linguistics}, 2024.

\bibitem[Zhao \& Tresp(2018)Zhao and Tresp]{zhao2018energy}
Zhao, R. and Tresp, V.
\newblock Energy-based hindsight experience prioritization.
\newblock In \emph{Conference on Robot Learning}, pp.\  113--122, 2018.

\bibitem[Zheng et~al.(2023)Zheng, Chiang, Sheng, Zhuang, Wu, Zhuang, Lin, Li, Li, Xing, Zhang, Gonzalez, and Stoica]{zheng2023judging}
Zheng, L., Chiang, W., Sheng, Y., Zhuang, S., Wu, Z., Zhuang, Y., Lin, Z., Li, Z., Li, D., Xing, E.~P., Zhang, H., Gonzalez, J.~E., and Stoica, I.
\newblock Judging llm-as-a-judge with mt-bench and chatbot arena.
\newblock In \emph{Advances in Neural Information Processing Systems}, 2023.

\bibitem[Zheng et~al.(2024)Zheng, Zhang, Zhang, Ye, Luo, Feng, and Ma]{zheng2024llamafactory}
Zheng, Y., Zhang, R., Zhang, J., Ye, Y., Luo, Z., Feng, Z., and Ma, Y.
\newblock Llamafactory: Unified efficient fine-tuning of 100+ language models.
\newblock \emph{arXiv preprint arXiv:2403.13372}, 2024.

\bibitem[Zhou et~al.(2023{\natexlab{a}})Zhou, Chen, Wan, Wen, Song, Yu, Huang, Peng, Yang, Xiao, et~al.]{zhou2023characterglm}
Zhou, J., Chen, Z., Wan, D., Wen, B., Song, Y., Yu, J., Huang, Y., Peng, L., Yang, J., Xiao, X., et~al.
\newblock Characterglm: Customizing chinese conversational ai characters with large language models.
\newblock \emph{arXiv preprint arXiv:2311.16832}, 2023{\natexlab{a}}.

\bibitem[Zhou et~al.(2023{\natexlab{b}})Zhou, Lu, Mishra, Brahma, Basu, Luan, Zhou, and Hou]{zhou2023instruction}
Zhou, J., Lu, T., Mishra, S., Brahma, S., Basu, S., Luan, Y., Zhou, D., and Hou, L.
\newblock Instruction-following evaluation for large language models.
\newblock \emph{arXiv preprint arXiv:2311.07911}, 2023{\natexlab{b}}.

\bibitem[Zhu et~al.(2025)Zhu, Huang, Lyu, Zhang, Li, Shi, Wu, Mu, Wang, Zhao, et~al.]{zhu2025codev}
Zhu, Y., Huang, D., Lyu, H., Zhang, X., Li, C., Shi, W., Wu, Y., Mu, J., Wang, J., Zhao, Y., et~al.
\newblock Codev-r1: Reasoning-enhanced verilog generation.
\newblock \emph{arXiv preprint arXiv:2505.24183}, 2025.

\end{thebibliography}
\bibliographystyle{icml2026}

\newpage
\appendix
\onecolumn
\part*{Appendix} 

\section{Implementation Details}
\label{app:implementation}
All experiments run on 8×A100-80GB GPUs. We use LLaMA-Factory~\citep{zheng2024llamafactory} for SFT and DPO training, and verl~\citep{sheng2025hybridflow} for RL training. The detailed training configurations of SFT and DPO are provided in Table~\ref{tab:config_sft}, and the training configurations of RL are provided in Table~\ref{tab:config_rl}.

\begin{table}[htbp]
\centering  
\vspace{-5pt}
\caption{Training configurations across different methods and model backbones.}
\begin{subtable}{\linewidth}
\subcaption{Training configuration of SFT and DPO.}
\label{tab:config_sft}
\begin{tabular}{@{}>{\bfseries}l p{12cm}@{}}
\toprule
Method & SFT, DPO \\
\midrule
Training
& per\_device\_train\_batch\_size = 16, gradient\_accumulation\_steps = 16 \\
& learning\_rate = 1e-6, lr\_scheduler\_type = constant \\
& cutoff\_len = 4096, warmup\_steps = 10, epochs = 5 \\
\midrule
Optimizations
& deepspeed: z3, bf16 \\
\bottomrule
\end{tabular}
\centering
\end{subtable}
\vspace{1em}
\begin{subtable}{\linewidth}
\vspace{5pt}
\subcaption{Training configuration of RL.}
\label{tab:config_rl}
\begin{tabular}{@{}>{\bfseries}l p{12cm}@{}}
\toprule
Method & RL-IR, RL-CR, HiR \\
\midrule
Sampling
& top\_k = -1, top\_p = 1.0, temperature = 1.0, rollout\_n = 8 \\
& max\_prompt\_length = 2,048, max\_response\_length = 4,096 \\
\midrule
Training
& ppo\_mini\_batch\_size = 64, ppo\_micro\_batch\_size\_per\_gpu = 8 \\
& log\_prob\_micro\_batch\_size\_per\_gpu = 8 \\
& learning\_rate = 1e-6, kl\_loss\_coef = 1e-4, epochs = 5 \\
\midrule
Optimizations
& param\_offload, flash\_attn, bf16 \\

\bottomrule
\end{tabular}
\centering
\end{subtable}
\vspace{-15pt}
\end{table}

We use the vLLM~\citep{kwon2025efficient} engine to generate responses for evaluation. The generation temperature is set to 0.6, and the maximum output length is set to 4,096 tokens. We report the average of five independent evaluation results across all benchmarks. For instruction following tasks, we use the default prompt template of models in evaluation. For OOD tasks (\textit{i.e.}, MATH-500, GPQA, MMLU-Pro), we add additional CoT prompts in evaluation as shown in Table~\ref{tab:eval_prompt}.

\begin{table}[htbp]
\vspace{-5pt}
\caption{Evaluation prompts on initial model across out-of-domain benchmarks.}
\vspace{-10pt}
\label{tab:eval_prompt}
\vspace{1pt}
\begin{center}
\begin{small}
\setlength{\tabcolsep}{6pt}
\renewcommand{\arraystretch}{1.2}
\resizebox{\textwidth}{!}{
\begin{tabular}{@{\extracolsep{\fill}}ll}
\toprule
\textbf{Dtasets} & \textbf{CoT Prompts} \\
\midrule
MATH-500 & Question: \{\}\texttt{\textbackslash n}Please reason step by step, and put your final answer within \texttt{\textbackslash boxed\{\}}. \\
\midrule
\multirow{3}{*}{GPQA \& MMLU-Pro} & Question: \{\}\texttt{\textbackslash n}Answer the multiple choice question. The last line of your response should \\
&be of the following format: 'Answer: \$LETTER' (without quotes) where LETTER is one \\
& of choices. Think step by step before answering. \\
\bottomrule
\end{tabular}%
}
\end{small}
\end{center}
\vspace{-10pt}
\end{table}

\section{Proof and Analysis}
\label{app:proof}
\subsection{Technical Settings and Notations}
\textbf{Settings}. Since the order of samples does not affect subsequent analysis, we assume that samples with indices from $i=1$ to $i=k$ are the failed samples used to replay for convenience. Our theoretical settings involve two main simplifications on Eq. (\ref{eq:hir}). First, we omit the clipping operation, because the clipping mechanism primarily serves as a practical stabilization heuristic to limit excessively large policy updates. Tokens that are out of range will not contribute to gradient, so the omission of the clipping does not affect the trajectory-level analysis. Second, we omit the nuanced differences in advantages among tokens within a response, as the KL coefficient is relatively small and will not be dominant.

\textbf{Notations}. We use $\pi_\theta$ to denote Large Language Models (LLMs) parameterized by $\theta$. The response $y^w$ and $y^l$ denote the winning (positive) and losing (negative) responses, and $y^r$ denotes the responses that are selected for replay.

\subsection{Proof of Proposition~\ref{pro:constract}}
\begin{proof}
The HiR training objective that omits the clipping mechanism is:
\begin{equation}
\begin{aligned}
\mathcal{J}_{\text{HiR}}(\theta) = & \mathbb{E}_{\substack{q \sim \mathcal{D} \\ \{y^{(i)}\}_{i=1}^{m} \sim \pi_{\text{old}}(\cdot|q) \\ \{q'^{(i)}, y'^{(i)}\}_{i=1}^{k} \sim \mathcal{H}}} \left[ \frac{1}{m} \sum_{i=1}^{m} \frac{1}{|y^{(i)}|} \sum_{t=1}^{|y^{(i)}|}  \rho_{t, \theta}^{(i)} A_{t}^{(i)} +
{\frac{1}{k} \sum_{i=1}^k \frac{1}{|y'^{(i)}|} \sum_{t=1}^{|y'^{(i)}|} \rho_{t, \theta}'^{(i)} A_{t}'^{(i)}} \right] \\
= & \mathbb{E}_{\substack{q \sim \mathcal{D} \\ \{y^{(i)}\}_{i=1}^{m} \sim \pi_{\text{old}}(\cdot|q) \\ \{q'^{(i)}, y'^{(i)}\}_{i=1}^{k} \sim \mathcal{H}}} 
 \left[ \frac{1}{m} \sum_{i=1}^{m} \frac{1}{|y^{(i)}|} \sum_{t=1}^{|y^{(i)}|}  \frac{\pi_\theta(y_{t}^{(i)} \mid q, y_{<t}^{(i)})}{\pi_{\text{old}}(y_{t}^{(i)} \mid q, y_{<t}^{(i)})} A_{t}^{(i)} +
{\frac{1}{k} \sum_{i=1}^k \frac{1}{|y'^{(i)}|} \sum_{t=1}^{|y'^{(i)}|} \frac{\pi_\theta(y_{t}'^{(i)} \mid q'^{(i)}, y_{<t}'^{(i)})}{\pi_{\text{old}}(y_{t}^{(i)} \mid q, y_{<t}^{(i)})} A_{t}'^{(i)}} \right].
\end{aligned}
\end{equation}
According to the standard importance sampling formula $\mathbb{E}_{q} \left[\frac{p(x)}{q(x)} f(x) \right] = \mathbb{E}_{p} [f(x)]$, we can obtain the objective as:
\begin{equation}
\begin{aligned}
\mathcal{J}_{\text{HiR}}(\theta) = & \mathbb{E}_{q \sim \mathcal{D}, \{y^{(i)}\}_{i=1}^{m} \sim \pi_{\theta}(\cdot|q), \{q'^{(i)}, y'^{(i)}\}_{i=1}^{k} \sim \mathcal{H}} \\
& \left[ \frac{1}{m} \sum_{i=1}^{m} \frac{1}{|y^{(i)}|} \sum_{t=1}^{|y^{(i)}|}  \pi_\theta(y_{t}^{(i)} \mid q, y_{<t}^{(i)}) A_{t}^{(i)} + {\frac{1}{k} \sum_{i=1}^k \frac{1}{|y'^{(i)}|} \sum_{t=1}^{|y'^{(i)}|} \pi_\theta(y_{t}'^{(i)} \mid q'^{(i)}, y_{<t}'^{(i)}) A_{t}'^{(i)}} \right].
\end{aligned}
\end{equation}
By separating the failed responses used for replay (\textit{i.e.}, indices from $1$ to $k$) from original samples and based on the fact that $\{y'^{(i)}\}_{i=1}^{k} = \{y^{(i)}\}_{i=1}^{k}$, we can derive:
\begin{equation}
\begin{aligned}
\mathcal{J}_{\text{HiR}}(\theta) = & \mathbb{E}_{q \sim \mathcal{D}, \{y^{(i)}\}_{i=1}^{m} \sim \pi_{\theta}(\cdot|q), \{q'^{(i)}\}_{i=1}^{k} \sim \mathcal{H}} \\
& \biggl[ \frac{1}{m-k} \sum_{i=k}^{m} \frac{1}{|y^{(i)}|} \sum_{t=1}^{|y^{(i)}|}  \pi_\theta(y_{t}^{(i)} \mid q, y_{<t}^{(i)}) A_{t}^{(i)} + \\
& \frac{1}{k} \sum_{i=1}^{k} \frac{1}{|y^{(i)}|} \sum_{t=1}^{|y^{(i)}|}  \pi_\theta(y_{t}^{(i)} \mid q, y_{<t}^{(i)}) A_{t}^{(i)} +
\frac{1}{k} \sum_{i=1}^{k} \frac{1}{|y^{(i)}|} \sum_{t=1}^{|y^{(i)}|}  \pi_\theta(y_{t}^{(i)} \mid q'^{(i)}, y_{<t}^{(i)}) A_{t}'^{(i)} \biggr].
\end{aligned}
\end{equation}
We further divide the first term into two groups: positive (winning) and negative (losing) samples, which obtains:
\begin{equation}
\begin{aligned}
\mathcal{J}_{\text{HiR}}(\theta) = & \mathbb{E}_{q \sim \mathcal{D}, \{y^{(i)}\}_{i=1}^{m} \sim \pi_{\theta}(\cdot|q), \{q'^{(i)}\}_{i=1}^{k} \sim \mathcal{H}} \\
& \biggl[ \frac{1}{m-k} \biggl( \sum_{i=k}^{G^-} \frac{1}{|y^{(i)}|} \sum_{t=1}^{|y^{(i)}|} A^{-} \pi_\theta(y_{t}^{(i)} \mid q, y_{<t}^{(i)}) + \sum_{i=G^-}^{m} \frac{1}{|y^{(i)}|} \sum_{t=1}^{|y^{(i)}|} A^{+} \pi_\theta(y_{t}^{(i)} \mid q, y_{<t}^{(i)}) \biggl) + \\
& \biggl( \frac{1}{k} \sum_{i=1}^{k} \frac{1}{|y^{(i)}|} \sum_{t=1}^{|y^{(i)}|} A^{-} \pi_\theta(y_{t}^{(i)} \mid q, y_{<t}^{(i)}) + \frac{1}{k} \sum_{i=1}^{k} \frac{1}{|y^{(i)}|} \sum_{t=1}^{|y^{(i)}|} A'^{+} \pi_\theta(y_{t}^{(i)} \mid q'^{(i)}, y_{<t}^{(i)}) \biggl) \biggr] \\
= & \mathbb{E}_{q \sim \mathcal{D}, \{y^{(i)}\}_{i=1}^{m} \sim \pi_{\theta}(\cdot|q), \{q'^{(i)}\}_{i=1}^{k} \sim \mathcal{H}} \\
& \biggl[\biggl(\alpha_1 \cdot \frac{1}{m - G^-} \sum_{i=G^-}^{m} \frac{1}{|y^{(i)}|} \sum_{t=1}^{|y^{(i)}|} \pi_\theta(y_{t}^{(i)} \mid q, y_{<t}^{(i)}) - \beta_1 \cdot \frac{1}{G^{-}- k} \sum_{i=k}^{G^-} \frac{1}{|y^{(i)}|} \sum_{t=1}^{|y^{(i)}|} \pi_\theta(y_{t}^{(i)} \mid q, y_{<t}^{(i)}) \biggl) + \\
& \biggl( \alpha_2 \cdot \frac{1}{k} \sum_{i=1}^{k} \frac{1}{|y^{(i)}|} \sum_{t=1}^{|y^{(i)}|} \pi_\theta(y_{t}^{(i)} \mid q'^{(i)}, y_{<t}^{(i)}) - \beta_2 \cdot \frac{1}{k} \sum_{i=1}^{k} \frac{1}{|y^{(i)}|} \sum_{t=1}^{|y^{(i)}|} \pi_\theta(y_{t}^{(i)} \mid q, y_{<t}^{(i)}) \biggl) \biggr],
\end{aligned}
\end{equation}
where $\alpha_1 = \frac{m-G^-}{m-k} A^{+}, \beta_1 = -\frac{G^- - k}{m-k} A^{-}, \alpha_2 = A'^{+}, \beta_2 = -A^{-}$. Note that they are all positive values as $A^+ >0$ and $A^-<0$.

By the law of large numbers $\lim_{N \to \infty} \frac{1}{N} \sum_{i=1}^{N} f(y) = \mathbb{E}_{y \in \pi_\theta} \ f(y)$, the empirical mean of finite samples converges to the true expectation as the sample size $N \to \infty$. We thus reformulate the empirical objective as the expected training objective: 
\begin{equation}
\begin{aligned}
\mathcal{J}_{\text{HiR}}(\theta) = & \mathbb{E}_{q \sim \mathcal{D}, q' \sim \mathcal{H}} \\
& \biggl[\biggl(\alpha_1 \cdot \underset{\substack{y^w \sim \pi_{\theta}(\cdot | q)}}{\mathbb{E}} \frac{1}{|y^{w}|} \sum_{t=1}^{|y^{w}|} \pi_\theta(y_{t}^{w} \mid q, y_{<t}^{w}) - \beta_1 \cdot \underset{\substack{y^l \sim \pi_{\theta}(\cdot | q)}}{\mathbb{E}} \frac{1}{|y^{l}|} \sum_{t=1}^{|y^{l}|} \pi_\theta(y_{t}^{l} \mid q, y_{<t}^{l}) \biggl) + \\
& \biggl( \alpha_2 \cdot \underset{\substack{y^r \sim \pi_{\theta}(\cdot | q)}}{\mathbb{E}} \frac{1}{|y^{r}|} \sum_{t=1}^{|y^{r}|} \pi_\theta(y_{t}^{r} \mid q', y_{<t}^{r}) - \beta_2 \cdot \underset{\substack{y^r \sim \pi_{\theta}(\cdot | q)}}{\mathbb{E}} \frac{1}{|y^{r}|} \sum_{t=1}^{|y^{r}|} \pi_\theta(y_{t}^{r} \mid q, y_{<t}^{r}) \biggl) \biggr].
\end{aligned}
\end{equation}
Therefore, the final expected training objective of HiR can be written as a form of preference learning on both the response- and instruction-level:
\begin{equation}
\begin{aligned}
J_{\text{HiR}}(\theta) = &{\mathbb{E}}_{q \sim \mathcal{D}, q' \sim \mathcal{H}} \\ & \biggl[ \biggr( \underbrace{\alpha_1 \underset{y^w \sim \pi_{\theta}(\cdot \mid q)}{\mathbb{E}} \pi_\theta(\textcolor{OliveGreen}{y^w} \mid q) - \beta_1 \underset{y^l \sim \pi_{\theta}(\cdot \mid q)}{\mathbb{E}} \pi_\theta(\textcolor{red}{y^l} \mid q)}_{\text{Response-level Preference}}  \biggr) + \biggr( \underbrace{\alpha_2 \underset{y^r \sim \pi_{\theta}(\cdot \mid q)}{\mathbb{E}} \pi_\theta(y^r \mid \textcolor{OliveGreen}{q'}) - \beta_2 \underset{y^r \sim \pi_{\theta}(\cdot \mid q)}{\mathbb{E}} \pi_\theta(y^r \mid \textcolor{red}{q})}_{\text{Instruction-level Preference}}  \biggr) \biggr]
\end{aligned}
\end{equation}
where $\pi_\theta(y \mid q) = \frac{1}{|y|} \sum_{t=1}^{|y|} \pi_\theta (y_t\mid y_{<t}, q)$.
\end{proof}

\section{Dataset Information}
\label{app:data_info}
\subsection{Training}
To facilitate hindsight rewriting, we construct 16,969 queries with decomposable constraints in different scenarios. Specifically, we collect public data from various sources, including MulDimIF~\citep{ye2025multi}, VerIF~\citep{peng2025verif}, IFTrain~\citep{Pyatkin2025GeneralizingVI}, and Chatbot Arena~\citep{zheng2023judging}. We first break down the atomic constraints in the instruction to form a constraint set $\mathcal{C}$ and then filter instructions with less than 5 atomic constraints. Finally, we obtained the \textbf{\texttt{HIR-16K}} dataset with 76,456 hard constraints and 46,536 soft constraints (a ratio of 1.6:1).

\subsection{Evaluation}

\textbf{IFEval}~\citep{zhou2023instruction} is a benchmark for evaluating the instruction following ability of LLMs, focusing on a set of verifiable instructions. The dataset comprises 25 types of verifiable instructions and 541 prompts, with each prompt including one or more verifiable instructions, such as word-count constraints and keyword occurrence requirements. 

\textbf{IFBench}~\citep{Pyatkin2025GeneralizingVI} is designed to evaluate the precise instruction following generalization of LLMs, aiming to test whether models can generalize to previously unseen instruction types. The dataset contains 58 new verifiable constraints with corresponding verification functions and 294 prompts, covering word-count limits, formatting requirements, counting, copying, and sentence/word/character manipulations. Each prompt may include one or more constraints.

\textbf{CFBench}~\citep{zhang2025cfbench} is a comprehensive Chinese benchmark comprising 1,000 carefully curated samples, covering over 200 real-world scenarios and more than 50 natural language processing tasks. Each sample contains multiple constraints organized into 10 primary categories and over 25 subcategories, with constraints seamlessly integrated into the original instructions and complex combinations carefully handled. The benchmark uses a multi-dimensional evaluation framework with requirement prioritization to assess performance from multiple perspectives.

\textbf{InfoBench}~\citep{qin2024infobench} comprises 500 diverse instructions and 2,250 decomposed questions across multiple constraint categories, designed to test and analyze the instruction following capabilities of LLMs systematically. The constraints involved in each instruction are categorized into five types: Content, Linguistic, Style, Format, and Number.

\textbf{ComplexBench}~\citep{wen2024benchmarking} is designed to evaluate the ability of LLMs to follow complex instructions under different compositions of constraints. The dataset is built upon a hierarchical taxonomy of 1,150 complex instructions, encompassing 4 constraint types, 19 constraint dimensions, and 4 composition types. 

\textbf{MulDimIF}~\citep{ye2025multi} is an instruction following benchmark built upon a multi-dimensional constraint framework. It covers three constraint patterns, four constraint categories, and four difficulty levels, comprising 1,200 code-verifiable instruction following test samples. MulDimIF enables systematic and fine-grained evaluation of large language models under diverse constraint forms and varying levels of complexity.

\textbf{FollowBench}~\citep{jiang2024followbench} is a multi-level, fine-grained instruction following benchmark for LLMs, designed to systematically evaluate their ability to understand and execute constraints in real-world instruction scenarios. The benchmark explicitly models constraint elements from user instructions, covering 5 types of fine-grained constraints: Content, Situation, Style, Format, and Example.

\textbf{MATH-500}~\citep{lightman2023let} dataset contains 500 high school–level math problems, covering 7 major areas such as precalculus, algebra, number theory, counting \& probability, geometry, intermediate algebra, and precalculus.

\textbf{GPQA}~\citep{rein2024gpqa} is a challenging scientific multiple-choice question dataset, primarily authored by experts in biology, physics, and chemistry, comprising 448 questions in the main set. The questions are carefully curated to ensure both high expertise and difficulty.

\textbf{MMLU-Pro}~\citep{mmlupro2024wang} is a benchmark for advanced multi-disciplinary language understanding and reasoning, designed to comprehensively evaluate LLMs on complex, multi-domain tasks. The dataset spans 14 disciplines, including mathematics, physics, chemistry, law, engineering, psychology, and health, comprising 12,032 questions. It particularly emphasizes high-difficulty problems that require reasoning, and the number of answer choices has been expanded from 4 in the original MMLU to 10 to increase distractor complexity and discriminative power.

\section{Evaluation Prompt}
\label{app:prompt}
We adopt the following prompt template for judging whether soft constraints are satisfied during RL training. 

\begin{tcolorbox}[
    colback=green!3!white,
    colframe=green!50!black,
    title=Soft Constraints Evaluation Prompt Template,
    fonttitle=\bfseries,
    boxrule=0.8pt,
    arc=2mm,
    breakable,
    label={box:scaffolding_template}
]

Based on the provided Input (if any) and Generated Text, judge whether the generated text fulfills the Criteria Item with either a YES or NO choice. Your selection should be based on your judgment as well as the following rules:\\

- YES: Select `YES' if the generated text entirely fulfills the condition specified in the Criteria Item. However, note that even minor inaccuracies exclude the text from receiving a 'YES' rating. As an illustration, consider a Criteria Item "Each sentence in the generated text uses a second person". If even one sentence does not use the second person, the answer should NOT be 'YES'. To qualify for a `YES' rating, the generated text must be entirely accurate and satisfy the criteria.\\

- NO: Opt for `NO' if the generated text fails to meet the criteria or provides no information that could be utilized to judge. For instance, the Criteria Item asks "Is the second sentence in the generated text a compound sentence?" and the generated text only has one sentence. It offers no relevant information to judge whether this criteria is met. Consequently, the answer should be `NO'.\\

\textbf{Input:}\\
\{\textcolor{blue}{input\_text}\}
\\
\textbf{Generated Text:}\\
\{\textcolor{blue}{generated\_text}\}
\\
\textbf{Criteria Item:}\\
\{\textcolor{blue}{criteria\_item}\}
\\

You only need to judge whether the generated text satisfiy the given Criteria Item and do NOT affect by other requirements in Input (if any). Return either a `YES' or `NO' choice without any additional text in your response.
\end{tcolorbox}


\end{document}